\def\paperID{2015}
\def\confName{CVPR}
\def\confYear{2025}

\def\paperTitle{\dsname: A Dataset and Method for LiDAR Surface Normal Estimation}

\def\authorBlock{
    Du\v{s}an Mali\'c$^{1,2}$ \qquad
    Christian Fruhwirth-Reisinger$^{1,2}$ \qquad
    Samuel Schulter$^{3,\dag}$ \qquad
    Horst Possegger$^{1,2}$ \\
    $^1$Christian Doppler Laboratory for Embedded Machine Learning \\
    $^2$Institute of Visual Computing, Graz University of Technology\\
    $^3$Amazon \\
    {\tt\small \{dusan.malic, reisinger, possegger\}@tugraz.at}
}

\newif\ifreview 
\newif\ifarxiv \newcommand{\arxiv}{\arxivtrue}
\newif\ifcamera 
\newif\ifrebuttal 

\arxiv %

\pdfoutput=1
\documentclass[10pt,twocolumn,letterpaper]{article}
\ifreview \usepackage[review]{cvpr} \fi
\ifarxiv \usepackage[pagenumbers]{cvpr} \fi
\ifrebuttal \usepackage[rebuttal]{cvpr} \fi
\ifcamera \usepackage{cvpr} \fi

\usepackage{graphicx}	
\usepackage{amsmath}	
\usepackage{amssymb}	
\usepackage{booktabs}
\usepackage{times}
\usepackage{microtype}
\usepackage{epsfig}
\usepackage{caption}
\usepackage{float}
\usepackage{placeins}
\usepackage{color, colortbl}
\usepackage{stfloats}
\usepackage{enumitem}
\usepackage{tabularx}
\usepackage{xstring}
\usepackage{multirow}
\usepackage{xspace}
\usepackage{url}
\usepackage{subcaption}
\usepackage{xcolor}
\usepackage[hang,flushmargin]{footmisc}
\usepackage{suffix}
\usepackage{siunitx}
\usepackage{mathtools}
\usepackage{pifont}
\usepackage{pgfplots}

\ifcamera \usepackage[accsupp]{axessibility} \fi

\newcommand{\nbf}[1]{{\noindent \textbf{#1.}}}
\WithSuffix\newcommand\nbf*[1]{\noindent \textbf{#1}}

\newcommand{\supp}{supplemental material\xspace}
\ifarxiv \renewcommand{\supp}{appendix\xspace} \fi

\definecolor{violetish}{RGB}{120, 20, 180}
\newcommand{\R}[1]{{%
    \textbf{%
        \ifstrequal{#1}{1}{\textcolor{violetish}{E2Y4}}{%
        \ifstrequal{#1}{2}{\textcolor{teal}{YRvi}}{%
        \ifstrequal{#1}{3}{\textcolor{cyan}{Kgnh}}{%
        \ifstrequal{#1}{4}{\textcolor{magenta}{R#1}}{%
                           \textcolor{red}{R#1}%
        }}}}%
    }%
}}

\newcommand{\dsname}{LiSu\xspace}

\newcommand{\sota}{state-of-the-art\xspace}

\newcommand{\ptv}{PTv3\xspace}

\newcommand{\bestresult}[1]{\text{\textbf{#1}}}
\newcommand{\secbresult}[1]{\underline{#1}}

\newcommand{\cmark}{\ding{51}}
\newcommand{\xmark}{\ding{55}}

\newcommand\blfootnote[1]{%
  \begingroup
  \renewcommand\thefootnote{}\footnote{#1}%
  \addtocounter{footnote}{-1}%
  \endgroup
}

\definecolor{colorOurs}{HTML}{8ecae6}
\definecolor{colorSHSNet}{HTML}{023047}
\definecolor{colorDuetal}{HTML}{e76f51}
\definecolor{colorGraphFit}{HTML}{ffb703}
\definecolor{colorPCPNet}{HTML}{8338ec}
\definecolor{colorPCA}{HTML}{780000}
\definecolor{colorJet}{HTML}{ff006e}
\definecolor{colorNeuralGF}{HTML}{008DD5}
\definecolor{colorCMGNet}{HTML}{F56476}
\definecolor{colorNGL}{HTML}{735D78}

\definecolor{colorPlot1}{HTML}{045992}
\definecolor{colorPlot2}{HTML}{db6000}
\definecolor{colorPlot3}{HTML}{108010}
\definecolor{colorPlot4}{HTML}{db9b5b}
\definecolor{colorPlot5}{HTML}{108010}

\usepackage{xr-hyper}

\makeatletter
\newcommand*{\addFileDependency}[1]{
  \typeout{(#1)}
  \@addtofilelist{#1}
  \IfFileExists{#1}{}{\typeout{No file #1.}}
}

\makeatother
\newcommand*{\myexternaldocument}[1]{
    \externaldocument{#1}
    \addFileDependency{#1.tex}
    \addFileDependency{#1.aux}
}

\definecolor{cvprblue}{rgb}{0.21,0.49,0.74}
\usepackage[pagebackref,breaklinks,colorlinks,allcolors=cvprblue]{hyperref}
\usepackage[capitalize]{cleveref}
\crefname{section}{Sec.}{Secs.}
\crefname{table}{Table}{Tables}
\crefname{figure}{Fig.}{Figs.}

\ifarxiv \crefname{appendix}{App.}{Apps.}
\else \crefname{appendix}{Suppl.}{Suppls.} \fi

\unless\ifarxiv \myexternaldocument{_supplementary} \fi

\usepackage{tikz}
\usetikzlibrary{positioning, calc, external}
\pgfplotsset{compat=1.18}

\begin{document}
\title{\paperTitle}
\author{\authorBlock}

\twocolumn[{%
\renewcommand\twocolumn[1][]{#1}%
\vspace{-12mm}
\maketitle
\vspace{-10mm}
\begin{center}
    \captionsetup{type=figure}
    \begin{subfigure}[b]{0.32\textwidth}
         \centering
         \includegraphics[width=\textwidth]{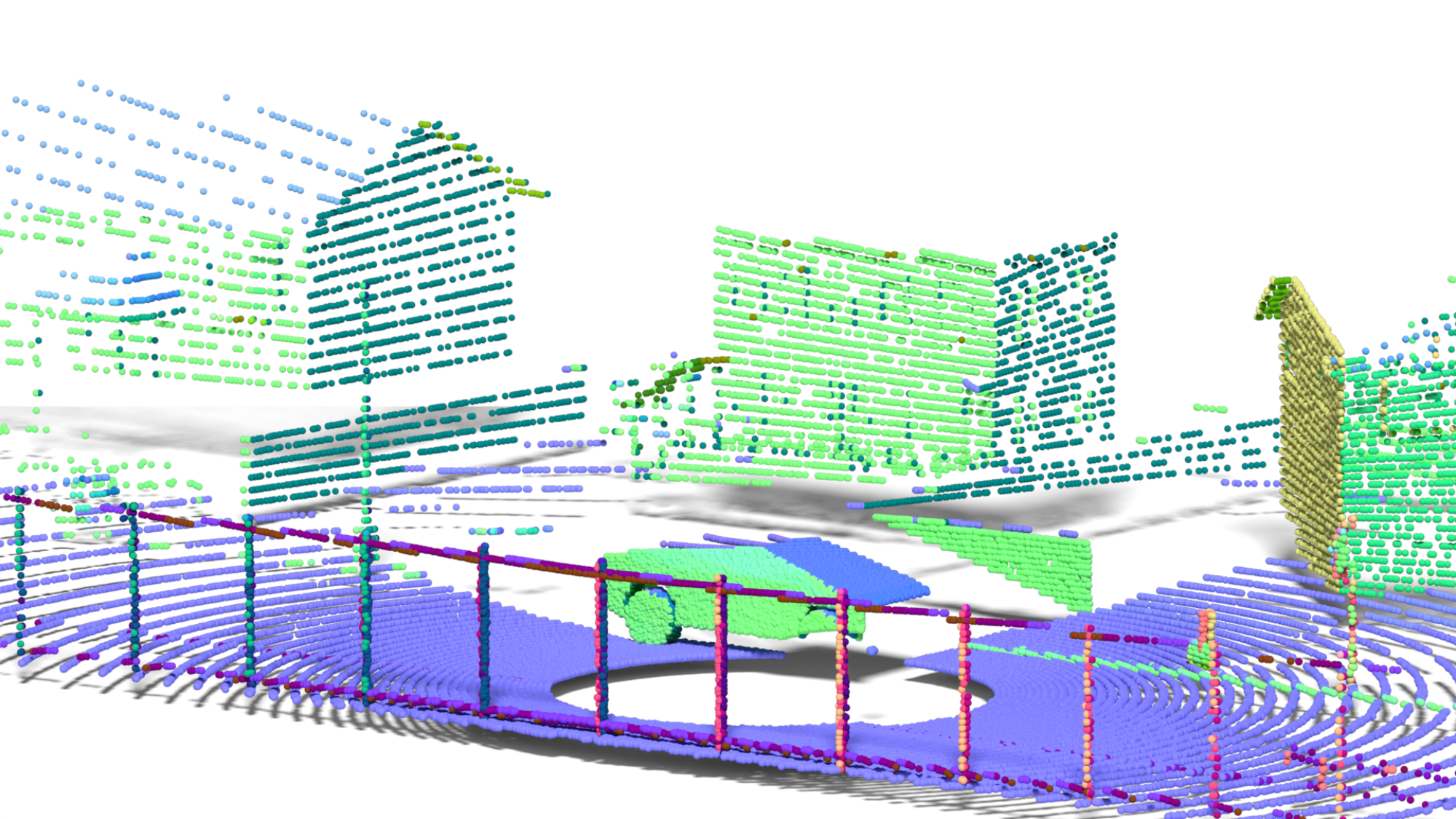}
         \caption{\dsname dataset sample}
         \label{fig:teaser_lisu}
     \end{subfigure}
    \begin{subfigure}[b]{0.32\textwidth}
         \centering
         \includegraphics[width=\textwidth]{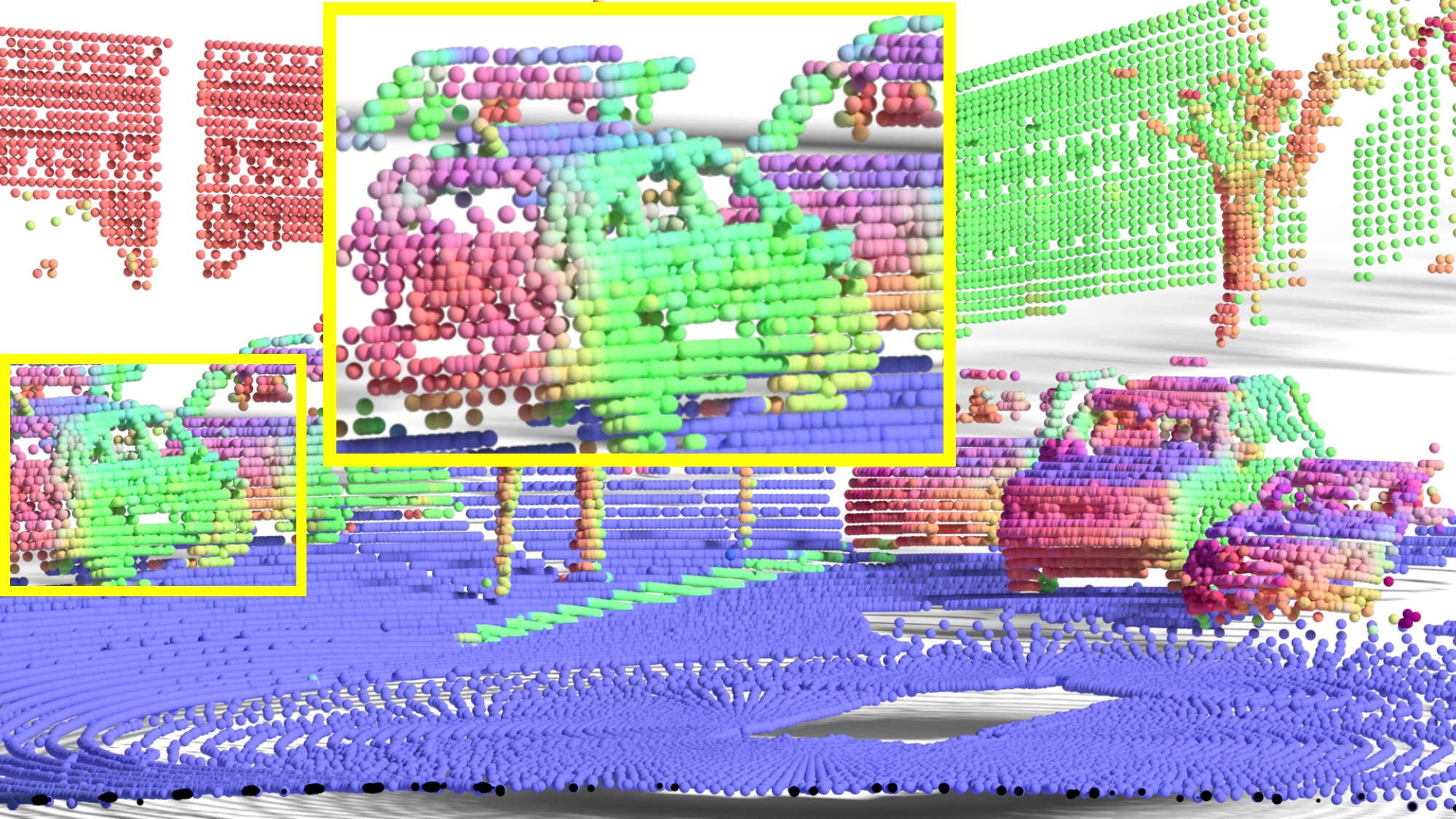}
         \caption{Our method applied on a Waymo frame}
         \label{fig:teaser_ours}
     \end{subfigure}
    \begin{subfigure}[b]{0.32\textwidth}
         \centering
         \includegraphics[width=\textwidth]{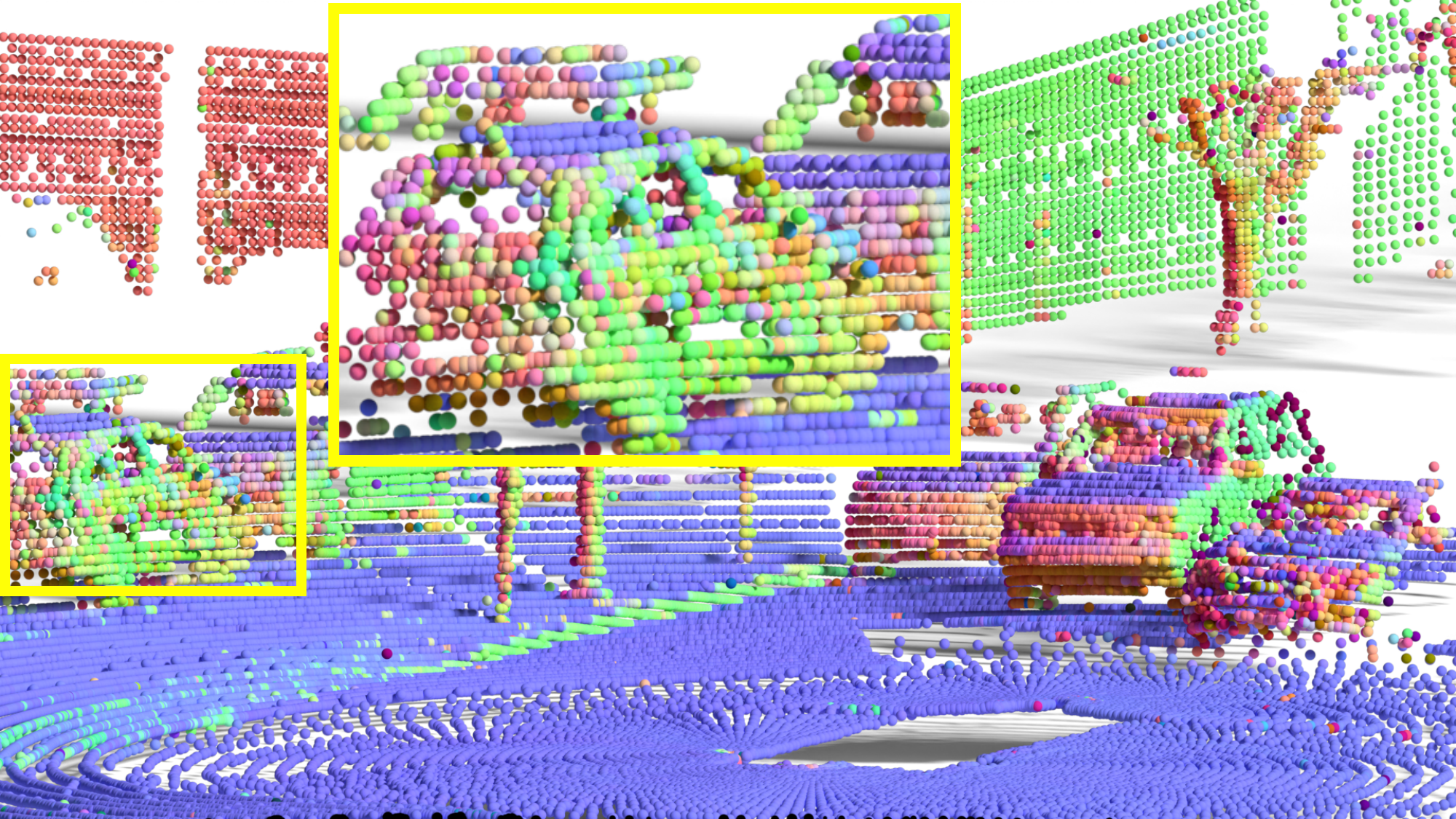}
         \caption{SHS-Net~\cite{liSHSNetLearningSigned2023} applied on a Waymo frame}
         \label{fig:teaser_shsnet}
     \end{subfigure}
    \setcounter{figure}{0}
    \setcounter{subfigure}{0}
     \captionof{figure}{
       Our synthetic \dsname dataset \hyperref[fig:teaser_lisu]{(a)} enables focusing research on the challenging task of LiDAR surface normal estimation.
      When combined with our proposed method, we achieve state-of-the-art results \hyperref[fig:teaser_lisu]{(b)} on challenging real-world datasets like Waymo Open Dataset~\cite{sunScalabilityPerceptionAutonomous2020}, outperforming the current state-of-the-art SHS-Net~\cite{liSHSNetLearningSigned2023} \hyperref[fig:teaser_lisu]{(c)}.
      Best viewed in color on screen.
    }
    \label{fig:teaser}
\end{center}%
}]
\blfootnote{$^\dag$Work conducted prior to joining Amazon}

\begin{abstract}
While surface normals are widely used to analyse 3D scene geometry, surface normal estimation from LiDAR point clouds remains severely underexplored.
This is caused by the lack of large-scale annotated datasets on the one hand, and lack of methods that can robustly handle the sparse and often noisy LiDAR data in a reasonable time on the other hand.
We address these limitations using a traffic simulation engine and present \dsname, the first large-scale, synthetic LiDAR point cloud dataset with ground truth surface normal annotations, eliminating the need for tedious manual labeling.
Additionally, we propose a novel method that exploits the spatiotemporal characteristics of autonomous driving data to enhance surface normal estimation accuracy.
By incorporating two regularization terms, we enforce spatial consistency among neighboring points and temporal smoothness across consecutive LiDAR frames.
These regularizers are particularly effective in self-training settings, where they mitigate the impact of noisy pseudo-labels, enabling robust real-world deployment.
We demonstrate the effectiveness of our method on \dsname, achieving state-of-the-art performance in LiDAR surface normal estimation. Moreover, we showcase its full potential in addressing the challenging task of synthetic-to-real domain adaptation, leading to improved neural surface reconstruction on real-world data.
\end{abstract}

\section{Introduction}
\label{sec:intro}

Representing 3D surfaces by their surface normals is beneficial for a wide range of computer vision tasks, such as neural rendering~\cite{verbinRefNeRFStructuredViewDependent2022, wangNeuralFieldsMeet2023, yuMonoSDFExploringMonocular2022}, robotics~\cite{behleyEfficientSurfelBasedSLAM2018, mur-artalORBSLAM2OpenSourceSLAM2017, schopsBADSLAMBundle2019, zhaiMonoGraspNet6DoFGrasping2023}, autonomous driving~\cite{miao3DObjectDetection2021, fanSNERoadSegIncorporatingSurface2020, tianGeoMAEMaskedGeometric2023}, \etc
While significant strides have been made in monocular surface normal estimation, particularly with large models trained on extensive datasets~\cite{baeRethinkingInductiveBiases2024, huMetric3DV2Versatile2024a, khirodkarSapiensFoundationHuman2024}, research on surface normal estimation from LiDAR point clouds in the autonomous driving domain remains limited.

Existing methods for point cloud surface normal estimation~\cite{duRethinkingApproximationError2023, liLearningSignedHyper2024, liNeuralGFUnsupervisedPoint2024, zhouRefinenetNormalRefinement2022} are tailored for small-scale CAD datasets, such as PCPNet~\cite{guerreroPCPNetLearningLocal2018} dataset, and often based on non-scalable architectures like PointNet~\cite{qiPointNetDeepLearning2017}.
Consequently, these methods struggle with large-scale point clouds and require either down-sampling or partitioning, which results in increased runtime.
They often assume dense and uniform point distributions, making them ill-suited for LiDAR data, which is characterized by sparsity, non-uniformity, and noise.
Moreover, the evaluation of these methods on LiDAR data is hindered by the lack of publicly available datasets with surface normal annotations.

To overcome these limitations, we present \dsname, a novel synthetic LiDAR dataset targeted for research on surface normal estimation.
We leverage CARLA~\cite{Dosovitskiy17}, a versatile simulation environment offering diverse urban and rural landscapes, including downtown areas, small towns, and multi-lane highways.
By extending CARLA's LiDAR sensor to capture not only point locations but also surface normal vectors, we curate an extensive dataset of roughly $50$k frames.

Leveraging the strong feature extraction capabilities of Point Transformer V3 (\ptv)~\cite{wuPointTransformerV32024}, we propose a single-step approach for surface normal estimation.
Our novel method explicitly accounts for the inherent noise in LiDAR data by incorporating two regularization terms: spatial consistency and temporal consistency.
These terms enforce smoothness and temporal coherence, respectively, leading to more robust and accurate normal estimates across consecutive frames.
Importantly, our method can be straightforwardly applied to domain adaptation scenarios, where our regularization terms effectively alleviate the adverse effects of noisy pseudo-labels in self-supervised learning~\cite{yangST3DDenoisedSelfTraining2023, zhangPseudoLabelRefinery2024, zhangReSimADZeroShot3D2024,yuanDensityguidedTranslatorBoosts2024, zhaoUniMixDomainAdaptive2024}.
This enables us to successfully bridge the synthetic-to-real domain gap, making our model suitable for real-world deployment.

We evaluate our method on our \dsname dataset, benchmarking it against two classical and four \sota methods for point cloud surface normal estimation.
In the absence of real-world datasets with surface normal annotations, we assess the effectiveness of our adaptation method by applying it to the real-world downstream task of neural surface reconstruction~\cite{guoStreetSurfExtendingMultiview2023, wangNeuRISNeuralReconstruction2022a, wangNeuSLearningNeural2021a, yuMonoSDFExploringMonocular2022}.
For this, we integrate our surface normal estimator model as an oracle into a LiDAR-only surface reconstruction method~\cite{zhangReSimADZeroShot3D2024}, demonstrating a substantial performance improvement.

In summary, our main contributions are:
\begin{itemize}
  \item Extending CARLA's LiDAR sensor to capture surface normals, enabling the generation of synthetic datasets for LiDAR-based surface normal estimation methods.
  \item \dsname: A novel, synthetic LiDAR dataset for autonomous driving, uniquely featuring labeled surface normals.
  \item Method employing data term regularizers to explicitly model spatiotemporal dependencies within autonomous driving datasets
  \item Extensive evaluation on both \dsname and real-world data.
  \item Publicly accessible code, dataset, and trained models at \url{https://github.com/malicd/LiSu}
\end{itemize}

\section{Related Work}
\label{sec:related}

\nbf*{Surface normal estimation from 3D point clouds}
involves the computation of the normal vector at each point on a 3D surface.
Traditional methods achieve this by fitting a geometric primitive, such as a tangent plane~\cite{hoppe1992surface}, jets~\cite{cazalsEstimatingDifferentialQuantities2005} or sphere~\cite{guennebaudAlgebraicPointSet2007}, to a local neighborhood of the query point.
The normal vector of this fitted primitive is then used as an estimate of the surface normal at the query point.
More recently, approaches leverage deep learning models to enhance surface fitting~\cite{ben-shabatDeepFit3DSurface2020, duRethinkingApproximationError2023, lenssenDeepIterativeSurface2020, liGraphFitLearningMultiscale2022, zhuAdaFitRethinkingLearningbased2021}.
Building upon the PointNet architecture~\cite{qiPointNetDeepLearning2017}, several methods~\cite{ben-shabatNestiNetNormalEstimation2019, guerreroPCPNetLearningLocal2018, liHSurfNetNormalEstimation2022} directly regress the surface normal vector from input point clouds.

These methods commonly utilize the PCPNet dataset~\cite{guerreroPCPNetLearningLocal2018}, which contains $30$ synthetic objects.
Each object consists of $100$k points representing geometric shapes and figurines.
To mitigate the limited data, Guerrero \etal~\cite{guerreroPCPNetLearningLocal2018} introduced a local patch sampling strategy, where $N$-point patches are extracted from the objects and used for training and inference.
However, this approach suffers from two limitations: patches from the same object share an underlying distribution, and complete shape estimation necessitates multiple inference steps.
The lack of data diversity and iterative inference routine render such approaches unsuitalbe for large-scale LiDAR point clouds.

\nbf*{Surface normal estimation from LiDAR point clouds}
remains a challenging problem due to the data's inherent sparsity, non-uniformity, and noise.
Traditional surface fitting methods often struggle with these characteristics.
Badino \etal~\cite{badinoFastAccurateComputation2011} addressed this by extending plane fitting to LiDAR range images and incorporating a normalization module for noise reduction.
More recent approaches, such as Bogoslavskyi \etal~\cite{bogoslavskyiFastRobustNormal2024}, leverage LiDAR scan lines to define local neighborhoods and infer surface normals.

With the rise of autonomous driving datasets (\eg \cite{geigerAreWeReady2012, sunScalabilityPerceptionAutonomous2020}), data from calibrated LiDAR and camera have become widely accessible.
This has spurred the development of multimodal methods like Lin \etal~\cite{linNormalTransformerExtracting2024} and DeepLiDAR~\cite{qiuDeepLiDARDeepSurface2019}, which leverage both modalities.
Scheuble \etal~\cite{scheublePolarizationWavefrontLidar2024} introduced PolLidar, a novel LiDAR sensor capable of capturing time-resolved polarimetric wavefronts.
They demostrate that surface normals can be derived from LiDAR characteristics beyond traditional geometric cues.
However, these methods rely on calibrated multi-sensor setups or specialized hardware, limiting their adoption in real-world applications.

The field of LiDAR surface normal estimation is hindered by the lack of a standardized benchmark dataset.
While the aforementioned works have made notable strides in addressing this, their datasets still exhibit limitations.
For example, DeepLiDAR primarily focuses on image-based tasks and employs LiDAR merely as sparse depth information, not providing LiDAR normals.
Additionally, the datasets of PolLidar~\cite{scheublePolarizationWavefrontLidar2024} and Lin \etal~\cite{linNormalTransformerExtracting2024} are relatively small ($1969$ and $151$ frames, respectively) and may not fully capture the diversity of real-world scenarios.
Crucially, these datasets were \emph{not publicly available} at the time of this submission.

To address these challenges, we introduce \dsname, a synthetic LiDAR dataset of $50$k frames with surface normal annotations, freely available to our research community.
Additionally, we propose a novel method to effectively leverage synthetic and unlabeled real-world data, enabling the development of robust models that generalize well to real-world scenarios.

\nbf*{Synthetic-to-Real Unsupervised Domain Adaptation (UDA) for LiDAR point clouds}
transfers knowledge from models trained on synthetic source data to real-world target datasets~\cite{xiaoTransferLearningSynthetic2022, yuanDensityguidedTranslatorBoosts2024, zhaoEPointDAEndEndSimulationReal2021, zhangReSimADZeroShot3D2024, zhangSRDANScaleawareRangeaware2021}.
Self-training is a widely employed UDA technique for various LiDAR-based tasks~\cite{micheleSALUDASurfacebasedAutomotive2024, saltoriGIPSOGeometricallyInformed2022, zhaoUniMixDomainAdaptive2024, chenRevisitingDomainAdaptive3D2023, yangST3DDenoisedSelfTraining2023, zhangPseudoLabelRefinery2024, zhangReSimADZeroShot3D2024}.
Our method closely aligns with existing UDA techniques.
However, we introduce a novel spatial and temporal regularization strategy to explicitly address the inherent noise in pseudo-labels.

\nbf*{Neural implicit surface reconstruction}
parametrizes 3D surfaces as Signed Distance Functions (SDFs) predicted by a multi-layer perceptron (MLP)~\cite{wangNeuSLearningNeural2021a, pengShapePointsDifferentiable2021, sitzmannImplicitNeuralRepresentations2020}.
A common optimization strategy involves guiding the learning objective with surface normal cues estimated with monocular off-the-shelf networks~\cite{groppImplicitGeometricRegularization2020a, guoStreetSurfExtendingMultiview2023, wangNeuRISNeuralReconstruction2022a, yuMonoSDFExploringMonocular2022}.
These approaches leverage the fact that SDF gradients at surface points correspond to inward surface normals~\cite{kazhdanPoissonSurfaceReconstruction2006}.
Naturally, the quality of normal estimates plays a crucial role~\cite{groppImplicitGeometricRegularization2020a}.
While LiDAR-only methods (\eg \cite{zhangReSimADZeroShot3D2024}) may neglect surface normal regularization, our work underscores the substantial benefits of precise surface normal estimations for superior surface reconstruction.

\section{Surface Normals from LiDAR Point Clouds}
\label{sec:sne_from_lidar}

To enable LiDAR-based surface normal estimation research, we introduce \dsname (\cref{sec:dataset}), a large-scale synthetic LiDAR dataset with surface normal annotations.
Moreover, we propose a novel method (\cref{sec:surf_estimation}) which leverages the spatiotemporal nature of autonomous driving data through two data term regularizers, achieving state-of-the-art performance on LiDAR surface normal estimation task.
This method also serves as a robust signal for mitigating pseudo-label noise in self-supervised learning.

\begin{table}[t]
\centering
\resizebox{\columnwidth}{!}{%

\begin{tabular}{lccccc}
\toprule
\textbf{Dataset}                                              & \textbf{\# Beams} & \textbf{PPF} & \textbf{\# Fr.} & \textbf{SN} & \textbf{Pub.} \\
\midrule
Waymo~\cite{sun2020wod}                                       & $\phantom{0}64$   & $160$k       & $230$k          & \xmark      & \cmark        \\
PolLidar~\cite{scheublePolarizationWavefrontLidar2024}        & $150$             & -            & $1.9$k          & \cmark      & \xmark        \\
Lin \textit{et al.}~\cite{linNormalTransformerExtracting2024} & -                 & -            & 151             & \cmark      & \xmark        \\
\dsname                                                       & $\phantom{0}64$   & $100$k       & $50$k           & \cmark      & \cmark        \\
\bottomrule
\end{tabular}
}
\caption{
  Overview of autonomous driving datasets, including LiDAR sensor configurations (number of beams, points per frame (PPF)), total frames (Fr.), surface normal (SN) annotation availability, and public (Pub.) accessibility.
}
\vspace{-0.05in}
\label{tab:dataset_info}
\end{table}

\subsection{\dsname: LiDAR Surface Normal Dataset}
\label{sec:dataset}
We generate our dataset using CARLA~\cite{Dosovitskiy17}, a simulation framework based on the Unreal Engine.
Specifically, we leverage nine of CARLA's twelve pre-built maps, excluding two reserved for the \emph{CARLA Autonomous Driving challenges} and one undecorated map with low geometric detail (\ie without buildings, sidewalks, \etc).
These selected maps represent diverse urban and rural environments, including downtown areas, small towns, and multi-lane highways.
For each simulation, we populated the scenes with a large number of dynamic actors, such as vehicles (cars, trucks, buses, vans, motorcycles, bicycles) and pedestrians (adults, children, police) as well as static props (barrels, garbage cans, road barriers, \etc).
The dynamic actors exhibited realistic movement patterns, governed by the underlying physics engine and adhered to real-world traffic rules, such as driving on designated roads and obeying traffic signals.

To capture realistic driving scenarios, we employ a virtual LiDAR sensor mounted atop a car operating in autopilot mode.
The LiDAR sensor is configured to emit $64$ laser beams, a $10^\circ$ upper and a $-30^\circ$ lower field of view.
Such a common sensor configuration (\eg \cite{geigerAreWeReady2012, sunScalabilityPerceptionAutonomous2020}) strikes a balance between sparsity and density, providing a challenging yet fair evaluation environment.
To further mimic real-world conditions, we set the maximum range to 100 meters and introduce Gaussian noise with a standard deviation of $0.02$ meters to the LiDAR point cloud.
The sensor captures data at a rate of \SI{10}{\hertz}.
We provide additional information about our simulation setup in the \supp.

\begin{figure}[tp]
    \centering
    \includegraphics[width=\linewidth]{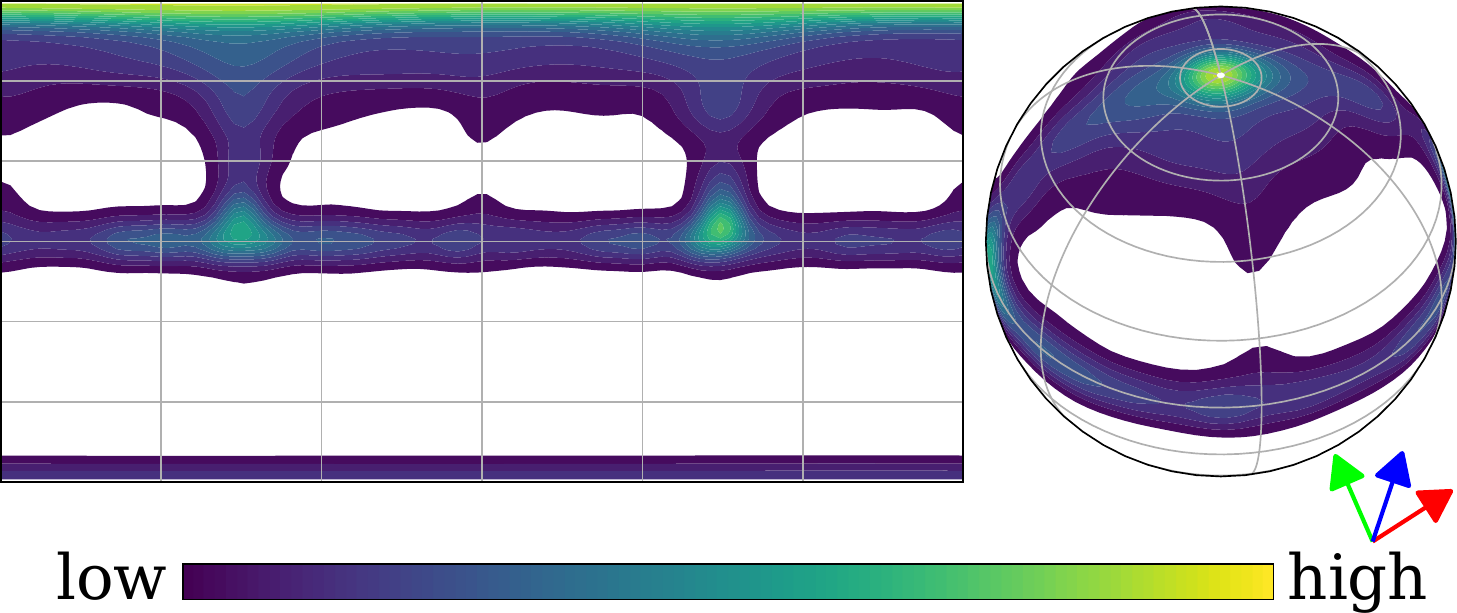}
    \caption{
      Spherical KDE plot using a von Mises-Fisher kernel to visualize surface normal distribution.
      Yellow regions indicate higher density of surface normals.
      White regions in the south correspond to physically impossible orientations, while those in the north represent extremely rare occurrences.
    }
    \label{fig:snorms_distribution}
\end{figure}

CARLA's default LiDAR sensor implementation is limited to position and intensity channels.
To enable surface normal collection, we extend CARLA's ray tracer to query surface normals at each intersection point between a ray and a mesh object.
These surface normals are then transformed into the sensor's coordinate frame and appended to the LiDAR data.
This requires modifications to both CARLA's C\texttt{++} backend and Python frontend, adding three extra channels to store the $x$, $y$, and $z$ components of the normal vector for each LiDAR point.

For each map, we conduct eleven randomly initialized and independent simulation runs.
A simulation is terminated early if prolonged traffic halts, such as red lights, occur.
On average, each simulation lasts approximately $50$ seconds, resulting in total of $\num{50045}$ labeled frames.
To ensure rigorous evaluation, we partition our dataset into training, validation, and testing sets.
We assign each map to exactly one split, preventing data leakage (\ie using the same ``city'' in multiple splits).
One map is designated for validation, while the remaining eight maps are divided equally between the training and testing sets.
This results in $\num{25053}$ training, $\num{22167}$ testing, and $\num{2825}$ validation frames.
The full dataset will be released publicly via a Research Data Management platform upon publication.

\begin{figure*}
    \centering
    \includegraphics[width=0.49\linewidth]{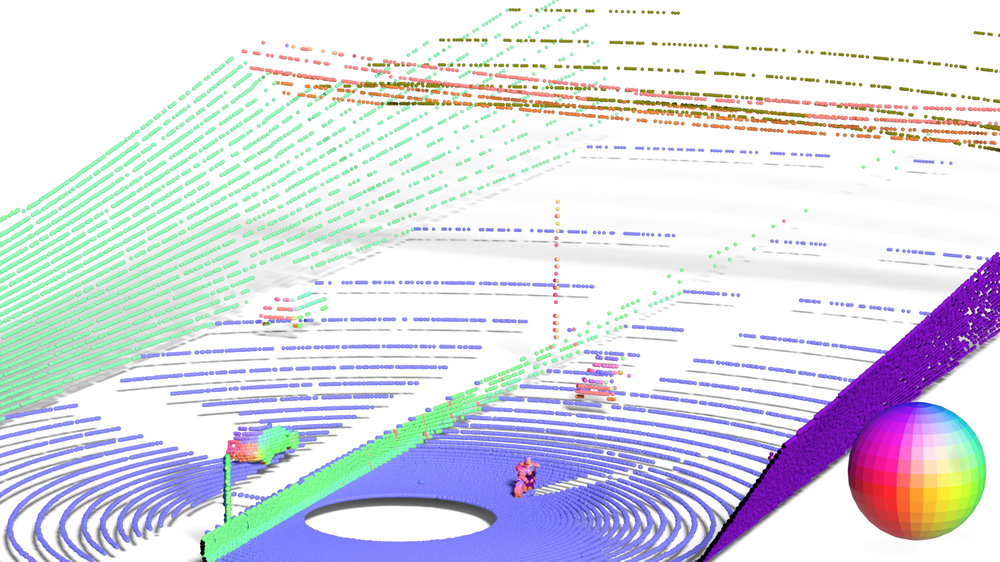}
    \hfill
    \includegraphics[width=0.49\linewidth]{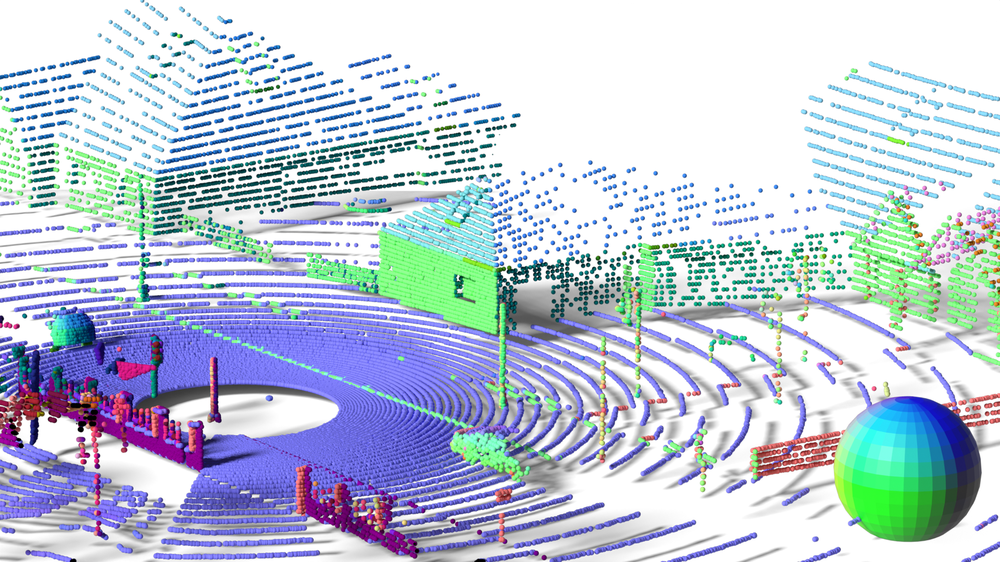}
    \caption{
      Exemplary LiDAR frames from our \dsname dataset (left: tunnel portal, right: urban scene).
      Surface normals are linearly mapped to the RGB color space.
      The color legend spheres in the bottom right corners provide a visual reference to interpret the normal directions.
    }
    \label{fig:dataset_examples}
\end{figure*}

Surface normals, being unit vectors, can be intuitively visualized as points on a unit sphere.
\cref{fig:snorms_distribution} illustrates a spherical Kernel Density Estimate (KDE) plot using a von Mises-Fisher kernel to represent the distribution of surface normals.
Our CARLA simulation, mirroring real-world environments, inherently produces imbalanced datasets.
For instance, the southern hemisphere, excluding a minor region around the south pole corresponding to ceiling surfaces (\eg in tunnels in \cref{fig:dataset_examples}), is largely unoccupied, since it is not physically possible for a surface to be oriented away from the LiDAR sensor.
Additionally, real-world scenes are dominated by road surfaces (concentrated near the north pole) and wall surfaces (near the equator), leading to a highly skewed dataset.
We address the challenges posed by this data imbalance in \cref{sec:ex_in_domain}.

\subsection{LiDAR Surface Normal Estimation} 
\label{sec:surf_estimation}

Given a LiDAR point cloud $\mathbf{P} \in \mathbb{R}^{N \times 3}$, we aim to learn a mapping $\Phi: \mathbb{R}^{N \times 3} \mapsto \mathbb{R}^{N \times 3}$ that, for each point $\mathbf{p}_i \in \mathbf{P}$, predicts a corresponding surface normal $\hat{\mathbf{n}}_i \in \mathbb{R}^3$.
While our method is architecture agnostic, we employ \sota Point Transformer V$3$ (PTv$3$)~\cite{wuPointTransformerV32024} as $\Phi$.
Given the ground truth label $\mathbf{n}$, our loss function is defined as
\begin{align}
  \mathcal{L}_{L1} = \frac{1}{N} \sum_{i \in N} || \mathbf{n}_i - \hat{\mathbf{n}}_i ||_1 \text{.}
  \label{eq:loss_l1}
\end{align}

While this type of supervision can guide the overall training procedure towards a local minima, it often fails to guarantee spatial and temporal coherence or accurate unit sphere estimates:
As a result, the model may produce inconsistent predictions, such as assigning opposing normal directions to points on the same surface or generating significantly different surface normal estimates for the same surface across consecutive frames.

To overcome these limitations, we propose three additional data terms that exploit the following key observations: (1) normal vectors should be piecewise smooth, (2) normal vectors should exhibit temporal consistency, and (3) normal vectors, being unit vectors, should be constrained to the unit sphere.
To enforce both spatial and temporal coherence, we adopt Graph Total Variation (GTV) regularization.
While our formulation shares similarities with prior works in point cloud super-resolution~\cite{dinesh3DPointCloud2019, dineshSuperResolution3DColor2020}, we deviate by incorporating GTV as a data term regularizer instead of the primary objective function.
In the following, we delve deeper into our regularization terms and present the overall loss objective.

\nbf{Spatial GTV (SGTV) Regularization}
The spatial regularization term operates under the assumption that geometrically close LiDAR points belong to the same surface and thus share the same surface normal vector.
We first convert our point cloud $\mathbf{P}$ into a $k$-neighborhood graph $\mathcal{G} = \left( \mathcal{V}, \mathcal{E}, \mathbf{W} \right)$, with set of nodes $\mathcal{V}$ and edges $\mathcal{E}$.
We associate each node $i \in \mathcal{V}$ with a location $\mathbf{p_i}$ and the respective surface normal prediction $\Phi \left( \mathbf{p}_i \right) = \hat{\mathbf{n}}_i$.
The edge weights are defined with an adjacency matrix $\mathbf{W}_{ij} = \left[w_{ij}\right]$, where
\begin{align}
  w_{ij} = \exp \left( - \frac{||\mathbf{p}_i - \mathbf{p}_j||^2_2}{\sigma^2} \right) %
  \label{eq:gtv_weight}
\end{align}
considers the spatial proximity between points $p_i$ and $p_j$.
It assumes values close to $1$ for nearby points and tends towards $0$ as the distance between them grows.
The parameter $\sigma$ determines the scale at which proximity is measured, influencing the rate of decay of the edge weights with increasing distance.
During optimization, we penalize spatially inconsistent surface normal predictions by
\begin{align}
  \mathcal{L}_\text{SGTV} = \frac{1}{|\mathcal{E}|} \sum_{\left(i, j\right) \in \mathcal{E}} w_{ij} || \hat{\mathbf{n}}_i - \hat{\mathbf{n}}_j ||_1 \text{.}
\end{align}

\nbf{Temporal GTV (TGTV) Regularization}
LiDAR point clouds are commonly captured sequentially at high frequency (\eg \SI{10}{\Hz} as in ONCE~\cite{maoOneMillionScenes2021}, WOD~\cite{sunScalabilityPerceptionAutonomous2020}, or Argoverse~\cite{wilsonArgoverse2Next2021}), resulting in substantial overlap between consecutive LiDAR scans.
Global registration modules, such as GPS coupled with IMU, allow for alignment of these frames to a common key frame.
This is particularly effective for static objects, while dynamic objects, such as moving vehicles, may exhibit slight misalignments.
However, in typical autonomous driving settings, the number of LiDAR points corresponding to static scene contents vastly outnumbers the number of points corresponding to dynamic objects.
We exploit the sequential nature of these datasets to enforce temporally coherent surface normal predictions.

During training, we employ mini-batches comprising pairs of consecutive point clouds, $(\mathbf{P}^t, \mathbf{P}^{t+1})$, along with their corresponding global poses (with respect to the key frame), $(\mathbf{T}^t, \mathbf{T}^{t+1}) \in \mathbb{R}^{4 \times 4}$.
Both point clouds undergo random affine transformations, $(\mathbf{A}^t, \mathbf{A}^{t+1}) \in \mathbb{R}^{4 \times 4}$, encompassing rotations, translations, flips, and other standard augmentations.
For improved readability, we adopt the convention $\mathbf{P} \triangleq \mathbf{A} \mathbf{T} \mathbf{P}$ throughout the paper.

After the forward pass $\Phi \left( \mathbf{P}^t \right)$ and $\Phi \left( \mathbf{P}^{t+1} \right)$, we construct a bipartite graph $\mathcal{G} = \left( \{ \mathcal{V}^t, \mathcal{V}^{t+1} \}, \mathcal{E}, \mathbf{W} \right)$, where each node $i \in \mathcal{V}^t$ is associated with the transformed key frame point $\left( \mathbf{T}^{t} \mathbf{A}^{t} \right)^{-1} \mathbf{p}_i^t$ and key frame surface normal prediction $\left( \mathbf{T}^{t} \mathbf{A}^{t} \right)^{-1} \hat{\mathbf{n}}_i^t$.
Similarly, each node $j \in \mathcal{V}^{t+1}$ is associated with the point $\left( \mathbf{T}^{t+1} \mathbf{A}^{t+1} \right)^{-1} \mathbf{p}_j^{t+1}$ and surface normal prediction $\left( \mathbf{T}^{t+1} \mathbf{A}^{t+1} \right)^{-1} \hat{\mathbf{n}}_j^{t+1}$.
Analogous to \cref{eq:gtv_weight}, the weight adjacency matrix is defined as
\begin{align}
    w_{ij} = \exp \left( - \frac{\left\lVert \left( \mathbf{T}^{t} \mathbf{A}^{t} \right)^{-1} \mathbf{p}_i^t - \left( \mathbf{T}^{t+1} \mathbf{A}^{t+1} \right)^{-1} \mathbf{p}_j^{t+1} \right\rVert^2_2}{\sigma^2} \right) \text{.}
  \label{eq:tgtv_weight}
\end{align}
Finally, we penalize incoherence between two consecutive frames with
\begin{align}
  \mathcal{L}_\text{TGTV} = \frac{1}{|\mathcal{E}|} \sum_{\left(i, j\right) \in \mathcal{E}} w_{ij} \| & \left( \mathbf{T}^{t} \mathbf{A}^{t} \right)^{-1} \hat{\mathbf{n}}_j^{t} - \notag \\
                                                                                        & \left( \mathbf{T}^{t+1} \mathbf{A}^{t+1} \right)^{-1} \hat{\mathbf{n}}_j^{t+1} \|_1 \text{,}
  \label{eq:tgtv}
\end{align}
which enforces temporal smoothness.

\nbf{Eikonal Regularization}
Recent studies~\cite{groppImplicitGeometricRegularization2020, guoStreetSurfExtendingMultiview2023, wangNeuSLearningNeural2021} have shown that enforcing unit-norm constraints on SDF gradients at object boundaries significantly improves model robustness.
Following this principle, we constrain our predicted surface normals to lie on the unit sphere:
\begin{align}
  \mathcal{L}_\text{Eikonal} = \frac{1}{N} \sum_{i \in N} \left( \left\lVert \hat{\mathbf{n}}_i \right\lVert_2 - 1 \right)^2 \text{.}
  \label{eq:eikonal}
\end{align}

\nbf{Training Objective}
Our training objective is a weighted combination of the loss and data regularization terms:
\begin{align}
    \mathcal{L} = \mathcal{L}_{L1} + \gamma \left( \mathcal{L}_\text{SGTV} + \mathcal{L}_\text{TGTV} + \mathcal{L}_\text{Eikonal} \right) \text{.}
    \label{eq:total_loss}
\end{align}
We empirically determined the optimal trade-off parameter $\gamma$ to be $0.1$.

\section{Experiments}
\label{sec:experiments}
In the following, we present a comprehensive experimental evaluation demonstrating the superior performance of our method on \dsname for LiDAR surface normal estimation (\cref{sec:ex_in_domain}).
We further highlight the positive impact of our \dsname on direct transfer to real-world datasets (\cref{sec:synth-to-real}) and self-supervised domain adaptation.
Finally, we show how our regularization terms mitigate the negative effects of noisy pseudo-labels (\cref{sec:synth-to-real}) in real-world neural surface reconstruction.

\subsection{Surface Normal Evaluation}
\label{sec:ex_in_domain}

\nbf{Dataset}
Due to the lack of LiDAR datasets with surface normal annotations, in the following we conduct our experiments on our \dsname dataset.
We train our models on the designated training split, comprising $\num{25053}$ labeled samples.
Previous methods, including PCPNet~\cite{guerreroPCPNetLearningLocal2018} and SHS-Net~\cite{liSHSNetLearningSigned2023}, partition point clouds into $N$-point patches, necessitating multiple inference steps for each frame.
These methods require on average over $10$ seconds per frame, making comprehensive evaluation time-consuming.
To enable a fair comparison within a reasonable timeframe, we downsampled our test set to every fifth frame, reducing the number of test frames to $\num{4433}$.
In contrast, our proposed method processes the entire point cloud in a single inference step. This allows us to process each frame in approximately $50$ milliseconds.

\nbf{Metric}
Previous works on LiDAR surface normal estimation, such as \cite{linNormalTransformerExtracting2024}, have adopted evaluation protocols inspired by PCPNet~\cite{guerreroPCPNetLearningLocal2018}.
However, these protocols, which involve progressively adding artificial noise, are not ideal for inherently noisy LiDAR data. 
Other methods, like \cite{scheublePolarizationWavefrontLidar2024}, impose overly strict error thresholds (\eg $3^\circ$), making them equally unsuitable.
Instead, we propose a more realistic evaluation protocol that accounts for the natural noise and sparsity of LiDAR data.
More specifically, we report the mean (\textbf{mean}), median (\textbf{median}) and root mean square (\textbf{RMSE}) angular error in degrees.
Additionally, we provide accuracy metrics for various angular error thresholds: $\mathbf{5.0^\circ}$, $\mathbf{7.5^\circ}$, $\mathbf{11.25^\circ}$, $\mathbf{22.5^\circ}$, and $\mathbf{30.0^\circ}$.
These metrics indicate the percentage of predictions with angular error below the respective threshold.

\nbf{Baseline}
First, we benchmark our method against two unsupervised surface normal estimation methods: PCA~\cite{hoppe1992surface} and Jet~\cite{cazalsEstimatingDifferentialQuantities2005}.
For both, we utilize a neighborhood size of $k=32$ points and orient normals towards a common viewpoint.
Subsequently, we train seven state-of-the-art supervised methods on the training split of our \dsname dataset: PCPNet~\cite{guerreroPCPNetLearningLocal2018}, CMG-Net~\cite{wuCMGNetRobustNormal2024}, GraphFit~\cite{liGraphFitLearningMultiscale2022}, Du \etal~\cite{duRethinkingApproximationError2023}, NGL~\cite{liNeuralGradientLearning2023}, NeuralGF~\cite{liNeuralGFUnsupervisedPoint2024}, and SHS-Net~\cite{liSHSNetLearningSigned2023}.
To adapt to the characteristics of LiDAR data, we modify the original implementations by reducing the number of points per patch from $500$ (within a $0.1$ meter radius) to $32$.
To accommodate the two orders of magnitude increase in scale of the training data, we have opted to reduce the number of training epochs.
This allows us to maintain the same total number of iterations, ensuring that the model is exposed to an equivalent amount of training data.
We provide more detailed information in the \supp.

\nbf{Implementation Details}
We adopt the default PTv3~\cite{wuPointTransformerV32024} configuration, which consists of a four-stage encoder-decoder architecture.
We optimize our model using AdamW~\cite{loshchilovDecoupledWeightDecay2019} with a maximum learning rate of $0.003$ and a weight decay of $0.005$.
The learning rate schedule comprises a short (1 epoch) warm-up phase followed by cosine annealing~\cite{loshchilovSGDRStochasticGradient2017}.
We train our model on 8 A100 GPUs using mixed precision with a batch size of $32$.
Standard point cloud augmentation techniques, such as rotation, flipping, and scaling, are utilized.
Training is terminated after $50$ epochs.

To address the class imbalance in our dataset (illustrated in \cref{fig:snorms_distribution}), we employ a weighted loss function:
Each data sample is weighted inversely proportional to the frequency of its ground truth surface normal occurrence.
This effectively upweights underrepresented classes, such as less common surface orientations, while downweighting overrepresented classes.
Consequently, the model is encouraged to focus more on the challenging, less frequent samples.

\begin{table*}
\centering
\resizebox{\textwidth}{!}{%

\begin{tabular}{lccccccccc}
\toprule
\textbf{Method}                                       & \textbf{mean} $\downarrow$   & \textbf{median} $\downarrow$ & \textbf{RMSE} $\downarrow$    & $\mathbf{5.0^\circ}$ $\uparrow$ & $\mathbf{7.5^\circ}$ $\uparrow$ & $\mathbf{11.25^\circ}$ $\uparrow$ & $\mathbf{22.5^\circ}$ $\uparrow$ & $\mathbf{30.0^\circ}$ $\uparrow$ & \textbf{Avg. Runtime}        \\
\midrule
PCA~\cite{hoppe1992surface}                           & 76.18                        & 29.10                        & 110.95                        & 44.95                           & 46.86                           & 48.45                             & 51.02                            & 52.09                            & \phantom{0}\bestresult{0.06} \\
Jet~\cite{cazalsEstimatingDifferentialQuantities2005} & 39.02                        & 25.11                        & \phantom{0}53.96              & 11.08                           & 20.75                           & 33.31                             & 50.47                            & 56.97                            & 14.55                        \\
PCPNet~\cite{guerreroPCPNetLearningLocal2018}         & \phantom{0}9.88              & \phantom{0}1.42              & \phantom{0}20.56              & 67.42                           & 72.41                           & 76.87                             & 84.75                            & 88.10                            & 14.31                        \\
CMG-Net~\cite{wuCMGNetRobustNormal2024}               & 10.05                        & \phantom{0}2.44              & \phantom{0}20.24              & 65.15                           & 71.09                           & 75.94                             & 78.40                            & 80.03                            & 40.01                        \\
GraphFit~\cite{liGraphFitLearningMultiscale2022}      & \phantom{0}8.98              & \phantom{0}0.93              & \phantom{0}19.57              & 73.49                           & 75.68                           & 77.42                             & 79.88                            & 80.80                            & 19.42                        \\
Du \etal~\cite{duRethinkingApproximationError2023}    & \phantom{0}8.81              & \phantom{0}0.93              & \phantom{0}19.54              & 73.53                           & 75.92                           & 77.75                             & 80.05                            & 80.88                            & 16.26                        \\
NGL~\cite{liNeuralGradientLearning2023}               & \phantom{0}9.88              & \phantom{0}1.21              & \phantom{0}18.43              & 77.39                           & 79.48                           & 81.44                             & 84.70                            & 86.24                            & 26.27                        \\
NeuralGF~\cite{liNeuralGFUnsupervisedPoint2024}       & \phantom{0}8.57              & \phantom{0}1.19              & \phantom{0}19.57              & 67.08                           & 71.42                           & 74.84                             & 79.48                            & 81.37                            & 17.58                        \\
SHS-Net~\cite{liSHSNetLearningSigned2023}             & \phantom{0}\secbresult{6.46} & \phantom{0}\secbresult{0.67} & \phantom{0}\secbresult{18.40} & \secbresult{79.66}              & \secbresult{83.57}              & \secbresult{86.77}                & \secbresult{90.79}               & \secbresult{92.06}               & 24.27                        \\
Ours                                                  & \phantom{0}\bestresult{6.30} & \phantom{0}\bestresult{0.43} & \phantom{0}\bestresult{17.58} & \bestresult{81.10}              & \bestresult{84.44}              & \bestresult{87.32}                & \bestresult{91.66}               & \bestresult{93.09}               & \phantom{0}\secbresult{0.09} \\
\bottomrule
\end{tabular}

}
\caption{
    Evaluation of surface normal estimation methods on the \dsname test split.
    Mean, median, and root mean square (RMSE) angular error in degrees, as well as accuracy for various angular error thresholds, and average runtime per LiDAR point cloud (in seconds) are reported.
    \bestresult{Best} and \secbresult{second-best} values are highlighted in bold and underlined, respectively.
 }
\vspace{-0.05in}
\label{tab:ex_in_domain}
\end{table*}

\nbf{Results}
\Cref{tab:ex_in_domain} presents the comparison of the surface normal estimation methods.
Traditional methods, such as PCA~\cite{hoppe1992surface} and Jet~\cite{cazalsEstimatingDifferentialQuantities2005}, are known to be sensitive to noise and prone to ambiguities in the orientation of their surface normal estimates.
When applied to sparse and non-uniform LiDAR data, as demonstrated in \cref{tab:ex_in_domain}, these methods exhibit significant limitations, often producing unreliable results.

Our proposed training method allows us to build a PTv3-based surface normal estimator that comfortably surpasses \sota supervised methods in both accuracy and speed.
Compared to PCPNet~\cite{guerreroPCPNetLearningLocal2018}, GraphFit~\cite{liGraphFitLearningMultiscale2022}, and Du \etal~\cite{duRethinkingApproximationError2023}, we achieve a significant reduction of over $2.5^\circ$ in mean average angular error and an $8\%$ increase in accuracy at the $5.0^\circ$ threshold.
Additionally, we outperform SHS-Net~\cite{liSHSNetLearningSigned2023} by reducing the median angular error by $0.24^\circ$ and improve the $5.0^\circ$ accuracy by $1.44^\circ$.
Furthermore, our model processes a single \dsname frame in just $90$ milliseconds on average, significantly outperforming \sota methods like SHS-Net, PCPNet, and GraphFit (which require over $14$ seconds).
This substantial speedup is attributed to our approach of processing the entire LiDAR point cloud at once, eliminating the need for partitioning and multiple inference steps, as required by \sota methods.

\nbf{Regularization Ablation Study}
To assess the individual contribution of each component proposed in \cref{sec:surf_estimation}, we conduct an ablation study shown in \cref{tab:component_ablation_on_synthetic}.
Each component, when trained from scratch on our full training split, outperforms our baseline model (trained with $\mathcal{L}_{L1}$ loss only).
The model incorporating all proposed losses demonstrates the best overall performance.
\begin{table}
\centering

\begin{tabular}{ccccccccccc}
\toprule
$\mathcal{L}_{L1}$ & $\mathcal{L}_{\text{SGTV}}$ & $\mathcal{L}_{\text{TGTV}}$ & $\mathcal{L}_{\text{Eikonal}}$ & \textbf{RMSE} $\downarrow$ & $\mathbf{5.0^\circ}$ $\uparrow$ \\
\midrule
\cmark             &                             &                             &                                & 19.82                      & 78.70                           \\
\cmark             & \cmark                      &                             &                                & 17.78                      & 81.06                           \\
\cmark             &                             & \cmark                      &                                & \secbresult{17.72}         & \secbresult{81.08}              \\
\cmark             &                             &                             & \cmark                         & 17.74                      & 80.65                           \\
\cmark             & \cmark                      & \cmark                      & \cmark                         & \bestresult{17.58}         & \bestresult{81.10}              \\
\bottomrule
\end{tabular}
\caption{
  Influence of proposed components on the in-domain model performance.
  Models are trained on the full \dsname training split and evaluated on a $20\%$ test split.
}
\vspace{-0.05in}
\label{tab:component_ablation_on_synthetic}
\end{table}

\subsection{Synthetic-to-Real Domain Adaptation}
\label{sec:synth-to-real}

In the following, we qualitatively assess the \sota in surface normal estimation on real-world LiDAR point clouds from the Waymo Open Dataset~\cite{sunScalabilityPerceptionAutonomous2020}.
We highlight the limitations of traditional methods and show that direct transfer from our \dsname dataset significantly reduces the synthetic-to-real domain gap compared to the existing PCPNet~\cite{guerreroPCPNetLearningLocal2018} dataset.
Furthermore, we demonstrate how our method can effectively bridge this gap.

\begin{figure}
    \centering
    \includegraphics[width=\linewidth]{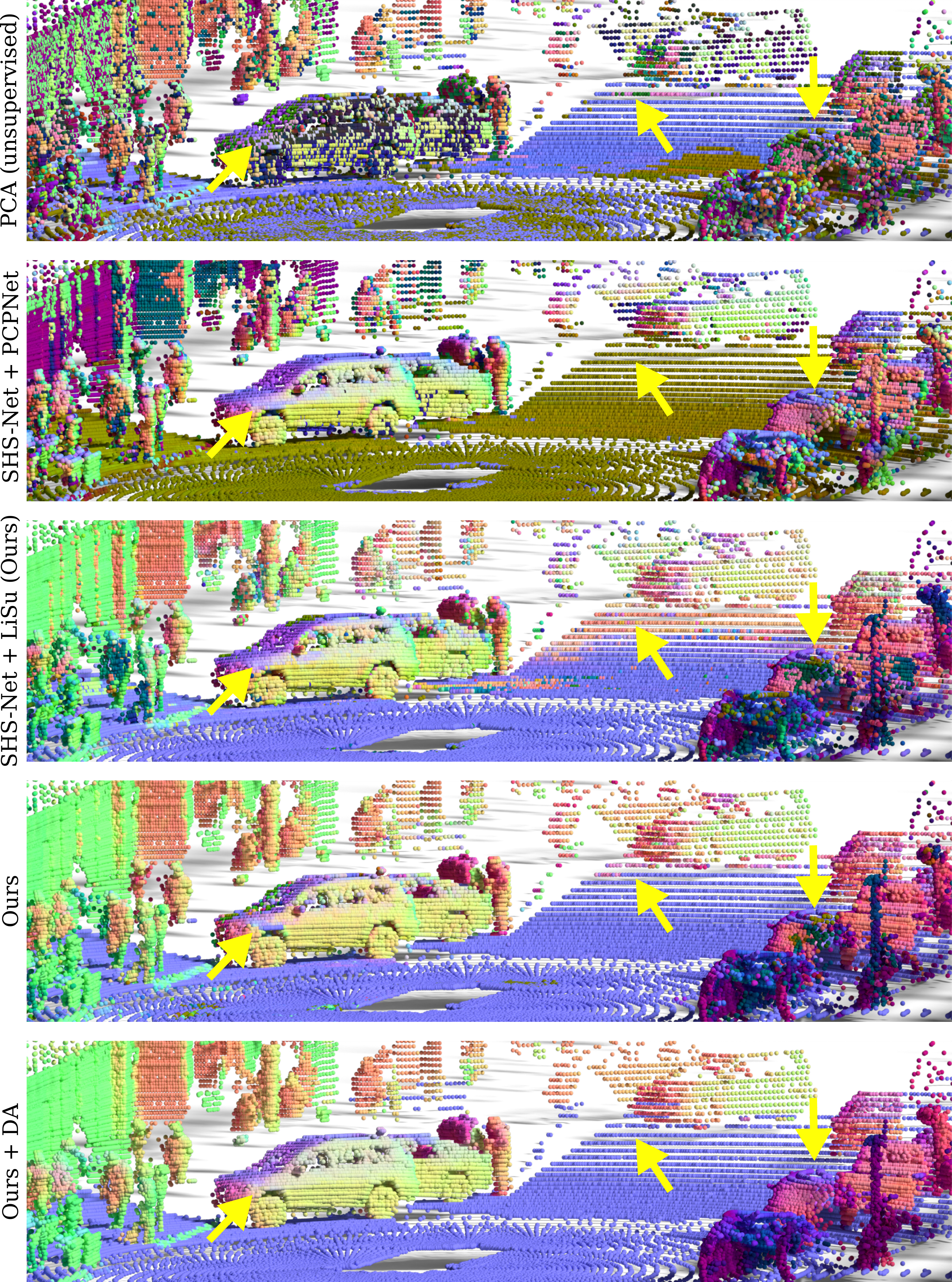}
    \caption{
      Qualitative evaluation of traditional and learning-based methods on a challenging Waymo frame.
      Notably, the current state-of-the-art (SHS-Net~\cite{liSHSNetLearningSigned2023} trained on PCPNet~\cite{guerreroPCPNetLearningLocal2018}) struggles to generalize to noisy and sparse LiDAR data.
      When trained on our proposed dataset (\dsname), both SHS-Net and our method yield reasonable results on the Waymo frame.
      In the last row, we demonstrate how leveraging our method within a self-supervised learning paradigm significantly enhances overall estimation, resulting in accurate, smooth, and consistently oriented predictions.
    }
    \label{figs:qwaymo}
\end{figure}

Classical approaches based on PCA~\cite{hoppe1992surface} (see Figure~\ref{figs:qwaymo}, first row) do not require labeled data but suffer from surface normal orientation ambiguities, particularly with noisy LiDAR point clouds.
For instance, points on a road with upward (blue) or downward (brown) normals can be interpreted as correct, depending on the perspective.
Yet, physical constraints dictate a single correct orientation, which is upward in this case.
More recent approaches, such as SHS-Net~\cite{liSHSNetLearningSigned2023} (second row in Figure~\ref{figs:qwaymo}), trained on the PCPNet dataset, exhibit similar limitations while providing smoother predictions.
Our direct domain transfer experiments (third and forth row in \cref{figs:qwaymo}), \ie models trained on our \dsname dataset and employed without any adaptation to Waymo, reveal already tremendous improvement.
This demonstrates that for the downstream task of LiDAR surface normal estimation, our \dsname dataset, as anticipated, is much better suited than existing CAD-based datasets like PCPNet.

To further enhance model robustness and generalization, we leverage our surface estimation method (\cref{sec:surf_estimation}) within a self-training paradigm, successfully reducing the domain gap.
We first pretrain our model on the \dsname training split, followed by a second phase on the Waymo train split, leveraging pseudo-labels and our proposed data term regularization to mitigate noise.
In the second (self-training) phase, we optimize our model using AdamW~\cite{loshchilovDecoupledWeightDecay2019} with a maximum learning rate of $0.001$ and weight decay of $0.005$.
To preserve the learned feature representation from the clean data, we employ differential learning rates~\cite{yangXLNetGeneralizedAutoregressive2019}, assigning lower learning rates to shallow layers and exponentially decaying them with network depth by a factor of $0.85$.
We employ standard point cloud augmentation techniques, such as rotation, flipping, and scaling.
The learning rate schedule employs a short warm-up phase followed by linear annealing~\cite{loshchilovSGDRStochasticGradient2017}. 
Our model is trained on 4 A100 GPUs using mixed precision with a batch size of $32$ for $10$ epochs.
In contrast to other approaches, our method consistently produces accurate, smooth, and correctly oriented surface normal predictions.

\subsection{Neural Surface Reconstruction}
\label{sec:ex_rel_world_data}
Our work demonstrates that the combination of our method with  \dsname effectively reduces the domain gap between datasets.
However, the specific influence of our regularization technique on the overall self-training process remains unclear.
Due to the scarcity of real-world LiDAR datasets with surface normal annotations, we opted to evaluate its effectiveness indirectly through a downstream task: neural surface reconstruction.

Previous methods in this domain often leverage geometric priors, such as surface normals derived from monocular images~\cite{groppImplicitGeometricRegularization2020a, guoStreetSurfExtendingMultiview2023, wangNeuRISNeuralReconstruction2022a}, to improve reconstruction quality.
These methods typically penalize discrepancies between the SDF boundary gradients and the predicted surface normals~\cite{yuMonoSDFExploringMonocular2022}.
However, LiDAR-only neural surface reconstruction approaches, such as ReSimAD~\cite{zhangReSimADZeroShot3D2024}, are limited by the absence of reliable surface normal estimation methods for real-world LiDAR data.
Therefore, we leverage our unsupervised training approach from \cref{sec:synth-to-real}, to learn a robust LiDAR surface normal estimator from the real-world Waymo dataset.
This prior is then integrated into ReSimAD to enhance its surface reconstruction capabilities.

We inherit the evaluation protocol established in \cite{zhangReSimADZeroShot3D2024, guoStreetSurfExtendingMultiview2023}:
For each sequence, we train a SDF model using point cloud data from all five Waymo LiDAR sensors, as described in \cite{zhangReSimADZeroShot3D2024}.
A single mesh is then reconstructed for the entire sequence using this trained model.
To simulate a real-world top-mounted LiDAR, virtual sensors are positioned at the locations of the real physical sensors and synthetic point clouds are generated by ray-casting the reconstructed mesh.
We evaluate the accuracy of these synthetic point clouds by comparing them to the corresponding real-world point clouds using Root Mean Square Error (RMSE) and Chamfer Distance (CD) metrics in \Cref{tab:implicit_surface_reconstruction}.

\begin{figure}
    \centering
    \includegraphics[width=\linewidth]{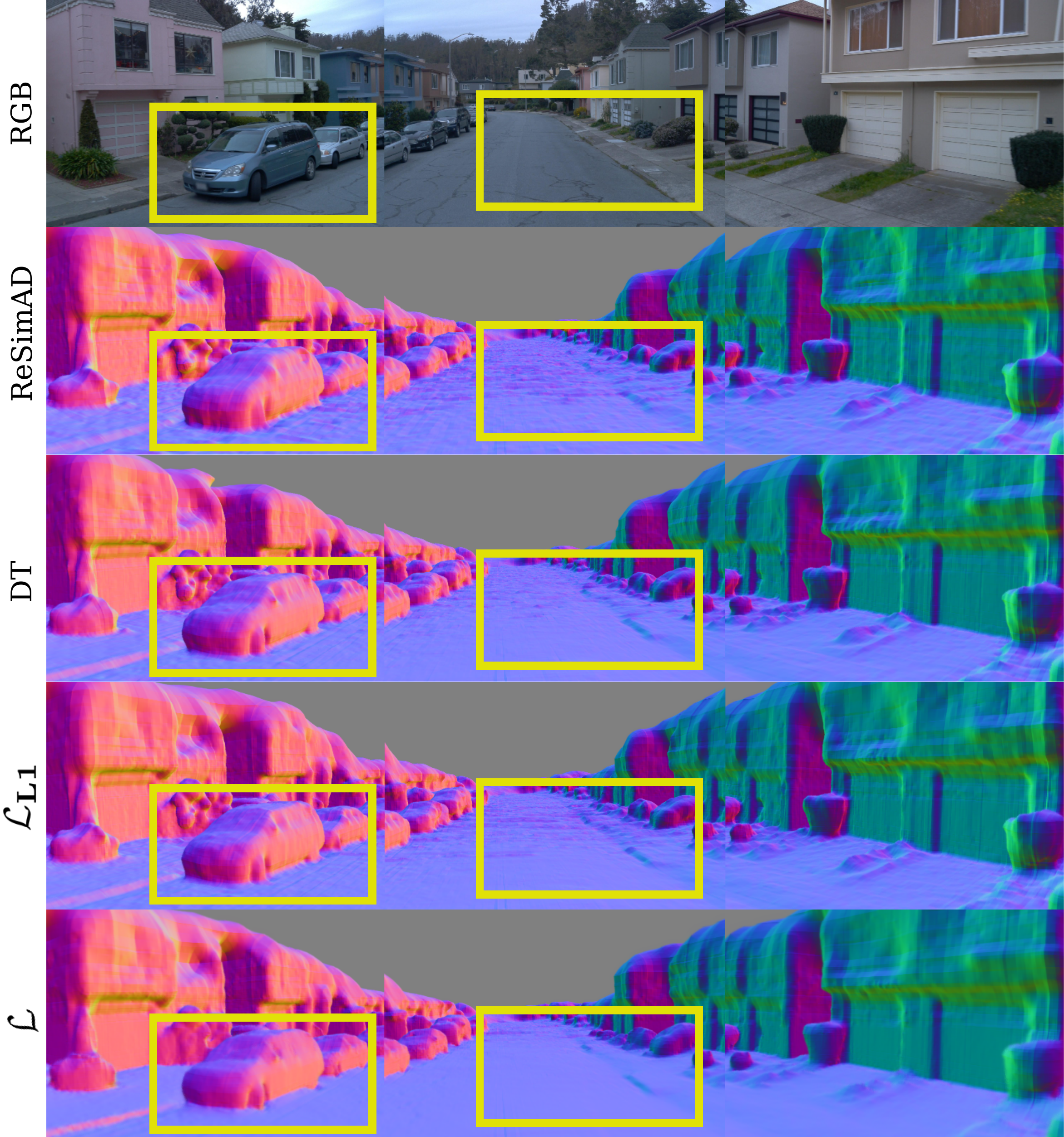}
    \caption{
      Reconstructed meshes from Waymo sequence 1172406, color-coded with surface normals.
      ReSimAD~\cite{zhangReSimADZeroShot3D2024} omits surface normal loss during reconstruction.
      Direct Transfer (DT) applies a model trained on \dsname without adaptation.
      $\mathbf{\mathcal{L}}$ (\cref{eq:total_loss}) and $\mathbf{\mathcal{L}_{L1}}$ (\cref{eq:loss_l1}) denote Waymo self-trained models with and without regularization, respectively.
    }
    \label{fig:neural_surface_reconstruction_ex}
\end{figure}

Surface reconstruction in ReSimAD is solely guided by the LiDAR point cloud information.
While this already produces reasonably good meshes, the SDF of regions that are not hit by a LiDAR beam remain undefined.
This leads to coarse meshes and jittering artifacts clearly visible in the exemplary ReSimAD results in \cref{fig:neural_surface_reconstruction_ex}.
Incorporating estimated surface normals into ReSimAD improves the mesh reconstruction notably, as illustrated in the three bottom rows of \cref{fig:neural_surface_reconstruction_ex}:
Directly transfering (DT) a surface normal estimator trained (solely) on \dsname to the Waymo data results in better surface normal estimates and, consequently, a smoother surface reconstruction near the ego vehicle.
Subsequently applying our proposed self-supervised domain adaption further improves the results: the bottom row of \cref{fig:neural_surface_reconstruction_ex} illustrates the benefits of our proposed regularization, which leads to smooth surface meshes, even at far distances from the ego vehicle.
Na\"ively performing self-supervised adaptation without our regularization (\ie solely using the $\mathcal{L}_{L1}$ loss from \cref{eq:loss_l1}), on the other hand, does not improve the results.

\begin{table}
\footnotesize
\centering
\begin{tabular}{ccccc}
\toprule
\textbf{Seq.} & \textbf{ReSimAD}~\cite{zhangReSimADZeroShot3D2024} & \textbf{DT}              & $\mathbf{\mathcal{L}_{L1}}$ & $\mathbf{\mathcal{L}}$                \\
\midrule
1006130       & 2.53~/~0.25                                        & \bestresult{2.51}~/~0.26 & \bestresult{2.51}~/~0.24    & \bestresult{2.51}~/~\bestresult{0.23} \\
1172406       & 1.95~/~0.52                                        & 1.95~/~0.50              & 1.96~/~0.51                 & \bestresult{1.90}~/~\bestresult{0.47} \\
1323841       & 2.16~/~0.70                                        & 2.15~/~0.75              & 2.16~/~0.74                 & \bestresult{2.12}~/~\bestresult{0.68} \\
1347637       & 2.07~/~0.33                                        & 2.05~/~0.25              & 2.05~/~0.29                 & \bestresult{2.02}~/~\bestresult{0.28} \\
1486973       & 1.80~/~\bestresult{0.19}                           & 1.83~/~0.21              & \bestresult{1.79}~/~0.20    & \bestresult{1.79}~/~0.20              \\
1506235       & 1.54~/~0.38                                        & 1.50~/~0.38              & 1.52~/~0.37                 & \bestresult{1.46}~/~\bestresult{0.34} \\
1664636       & 2.15~/~0.53                                        & 2.19~/~0.51              & 2.10~/~0.48                 & \bestresult{2.08}~/~\bestresult{0.47} \\
4058410       & 1.91~/~0.36                                        & 1.91~/~0.34              & 1.91~/~0.40                 & \bestresult{1.90}~/~\bestresult{0.31} \\
\midrule
average       & 2.01~/~0.41                                        & 2.01~/~0.40              & 2.00~/~0.40                 & \bestresult{1.97}~/~\bestresult{0.37} \\
\bottomrule
\end{tabular}
\caption{
    We evaluate neural surface reconstruction on diverse Waymo sequences using Root Mean Square Error (RMSE)~/~Chamfer Distance (CD).
    Lower values indicate better performance.
    ReSimAD~\cite{zhangReSimADZeroShot3D2024} omits surface normal loss during reconstruction.
    Direct Transfer (DT) applies a model trained on \dsname without adaptation.
    $\mathbf{\mathcal{L}}$ (\cref{eq:total_loss}) and $\mathbf{\mathcal{L}_{L1}}$ (\cref{eq:loss_l1}) denote Waymo self-trained models with and without regularization, respectively.
}
\vspace{-0.05in}
\label{tab:implicit_surface_reconstruction}
\end{table}

\section{Conclusion}
\label{sec:conclusion}
We introduced \dsname, a synthetic dataset designed to advance research on LiDAR-based surface normal estimation.
With more than \num{50000} LiDAR frames, obtained from a state-of-the-art traffic simulation engine, \dsname presents an ideal testbed to develop and benchmark LiDAR surface normal estimation approaches.
To properly leverage this large-scale synthetic data we propose a method that exploits the spatiotemporal nature of LiDAR scans in autonomous driving settings via meticulously designed regularization constraints.
Moreover, we show that our method can be seamlessly extended to domain adaptation scenarios, where it helps to alleviate the negative effects of noisy pseudo-labels.
Across several experiments, we demonstrate the utility of \dsname and our method to obtain outstanding results in surface normal estimation and their positive impact on the downstream task of surface reconstruction from real-world LiDAR point clouds.
We believe the \dsname dataset will help our research community to improve LiDAR-based surface normal estimation approaches and, consequently, leverage these to advance autonomous driving:
For example, scenario generation for domain adaptation research (\ie, re-rendering scenes with different simulated LiDAR configurations) will benefit from improved surface reconstructions.

\nbf{Acknowledgments} We gratefully acknowledge the financial support by the Austrian Federal Ministry for Digital and Economic Affairs, the National Foundation for Research, Technology and Development and the Christian Doppler Research Association. The presented experiments have been achieved using the Vienna Scientific Cluster.

{\small
\bibliographystyle{ieeenat_fullname}
\bibliography{11_references}
}

\ifarxiv \clearpage \appendix \maketitlesupplementary 
This supplementary material provides an in-depth analysis of inference speed for the models used in our benchmarks (\cref{sec:runtime_vs_acc}). Additionally, it offers comprehensive details on the acquisition of our \dsname dataset (\cref{sec:dataset_acquisition}) and the implementation specifics of baseline methods (\cref{sec:implementation_details_others}).
Furthermore, we present additional experiments for the neural surface reconstruction downstream task (\cref{sec:abl_neural_surface_reconstruction}), qualitative evaluations (\cref{sec:supp_qualitative}), and a rigorous ablation study exploring the impact of various design choices (\cref{sec:loss_reg_tradeoff}, \cref{sec:k_nbhood_graph_abl}, \cref{sec:gamma_ablation}).

\section{\texorpdfstring{Inference Speed \vs Accuracy}{Inference Speed vs. Accuracy}}
\label{sec:runtime_vs_acc}

Traditional methods like PCA~\cite{hoppe1992surface} are renowned for their efficient runtime.
However, their accuracy is degraded by inherent rotation ambiguities.
This often manifests as points on the same plane exhibiting opposite surface normal directions (recall Fig. 4 of the main manuscript). %
Common heuristics, such as orienting all normals towards a fixed viewpoint or propagating orientation information via Minimum Spanning Tree (MST)~\cite{hoppe1992surface}, alleviate this problem but, due to the noisy nature of LiDAR point clouds, are not particularly effective.

Supervised methods like SHS-Net~\cite{liSHSNetLearningSigned2023}, Du \etal~\cite{duRethinkingApproximationError2023}, GraphFit~\cite{liGraphFitLearningMultiscale2022}, and PCPNet~\cite{guerreroPCPNetLearningLocal2018} offer substantial performance gains, but at the cost of significant computational overhead.
A key limitation of these methods is their reliance on point cloud partitioning, required during both training and inference.
Furthermore, they often employ point-based backbone architectures like PointNet~\cite{qiPointNetDeepLearning2017} or architectures which require special operations such as DGCNN~\cite{wangDynamicGraphCNN2019}, which hinder efficient processing.
Originally introduced to address the limited dataset size of PCPNet~\cite{guerreroPCPNetLearningLocal2018} ($30$ samples), point cloud partitioning remains necessary during inference, significantly increasing processing time, especially for large-scale datasets like ours, \dsname (approximately $100$k points per frame).
Batching partitions is a potential strategy for accelerating inference.
However, GPU VRAM limits batch sizes, preventing single-frame inference and necessitating multiple inference passes for batched partitions.
Moreover, PointNet and DGCNN are not optimized for large-scale point clouds, making it challenging to adapt them to single-frame training/inference.

Conversely, we leverage the Point Transformer V3 (PTv3)~\cite{wuPointTransformerV32024}, a state-of-the-art transformer architecture build for large-scale LiDAR point clouds.
Their employment of space-filling curves (\eg z-order or Hilbert curve) for point cloud serialization and hardware-optimized operations (\eg FlashAttention~\cite{daoFlashAttention2FasterAttention2024, daoFlashAttentionFastMemoryefficient2022}) enable significant speedups.
A single inference step on a large-scale LiDAR point cloud can be executed in orders of magnitude less time (\eg $50$ milliseconds vs. $20$ seconds for DGCNN).
PTv3 coupled with our novel \dsname and training method exhibits exceptional performance in LiDAR surface normal estimation, requiring significantly less computational time compared to existing methods, as visualized in \cref{fig:runtime_accuracy_plot}.

\begin{figure}
\centering
\includegraphics{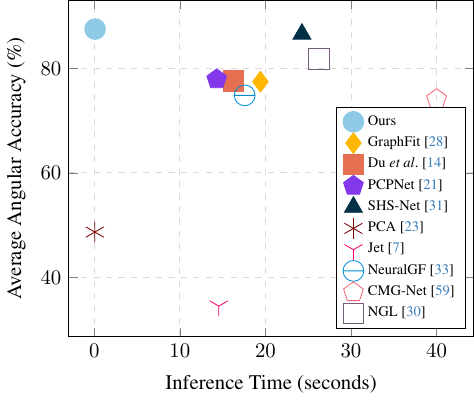}
\caption{
  Inference speed \vs accuracy plot for various neural network-based and traditional methods.
  Accuracy is calculated as the average angular accuracy across all thresholds listed in Tab. 2 of the main manuscript: $\{5.0^\circ, 7.5^\circ, 11.25^\circ, 22.5^\circ, 30.0^\circ\}$.
}
\label{fig:runtime_accuracy_plot}
\end{figure}

\section{\dsname Acquisition}
\label{sec:dataset_acquisition}

To obtain our custom \dsname dataset, we extended the CARLA simulator (version 0.9.15\footnote{\url{https://github.com/carla-simulator/carla/releases/tag/0.9.15}}) built with Unreal Engine 4.26\footnote{\url{https://github.com/CarlaUnreal/UnrealEngine}} with an additional LiDAR sensor.
This sensor captures not only standard position and intensity data but also the surface normal vector of each hit point.
We extended CARLA's ray caster to return surface normals in the sensor's frame of reference and stream this data from the C\texttt{++} core.
This data stream was collect by the Python front-end and saved to a file together with the global sensor position and orientation (required for keyframe transformation in TGTV in Sec.~3.2 of the main manuscript)

We initialize a virtual LiDAR sensor to reflect commonly employed real-world LiDARs~\cite{geigerVisionMeetsRobotics2013, sunScalabilityPerceptionAutonomous2020}.
A detailed configuration of this sensor is presented in Table~\ref{tab:lidar_sensor_config}.
The virtual LiDAR sensor was mounted on a self-driving car, randomly placed within the simulation environment.
We populated the simulation environment with approximately $\num{6000}$ dynamic actors, such as vehicles (cars, trucks, buses, vans, motorcycles, bicycles) and pedestrians (adults, children, police) as well as $\num{2000}$ static props (barrels, garbage cans, road barriers, \etc).
Due to the different map sizes (\eg \cref{fig:carla_maps_bev_town02} \vs \cref{fig:carla_maps_bev_town03}), not all objects were guaranteed to appear in every simulation.
For each map, we conducted $N$ independent runs, each initialized with a different random seed.
To avoid redundant frames, we terminated simulations prematurely if prolonged traffic halts, such as those caused by red lights, occurred.
A detailed breakdown of simulation runs is provided in \cref{tab:lisu_scenes}.

\begin{table}
\centering
\begin{tabular}{lc}
\toprule
\textbf{Description} & \textbf{Value} \\
\midrule
Number of lasers & $64$ \\
Maximum distance to raycast in meters & $100$ \\
Points generated by all lasers per second & $2$M \\
LIDAR rotation frequency & \SI{10}{\hertz} \\
Angle of the highest beam & $10^\circ$ \\
Angle of the lowest beam & $-30^\circ$ \\
Horizontal field of view & $360^\circ$ \\
Proportion of randomly dropped points & $0.45$ \\
Std Dev point noise along the raycast vector & $0.02$ \\
\bottomrule
\end{tabular}
\caption{
  Virtual LiDAR attributes used for the data acquisition.
}
\vspace{-0.05in}
\label{tab:lidar_sensor_config}
\end{table}

\begin{table*}
\centering
\resizebox{\textwidth}{!}{%

\begin{tabular}{ccccl}
\toprule
                                                                   & \textbf{Map}     & \textbf{\# Runs}  & \textbf{\# Frames}    & \textbf{Summary}                                                                                          \\
\midrule
  \multirow{4}{*}{\rotatebox[origin=c]{90}{\textit{train}}}        & Town01  & \num{11} & \phantom{0}\num{6339}   & \small A small, simple town with a river and several bridges.                                    \\
                                                                   & Town03  & \num{11} & \phantom{0}\num{7658}   & \small A larger, urban map with a roundabout and large junctions.                                \\
                                                                   & Town05  & \num{11} & \phantom{0}\num{5477}   & \small Squared-grid town with cross junctions and a bridge. It has multiple lanes per direction. \\
                                                                   & Town07  & \num{11} & \phantom{0}\num{5579}   & \small A rural environment with narrow roads, corn, barns and hardly any traffic lights.         \\
      \midrule
                                                             total & 4       & \num{44} & \num{25053}  &                                                                                                  \\
      \midrule
   \multirow{4}{*}{\rotatebox[origin=c]{90}{\textit{test}}}        & Town02  & \num{11} & \phantom{0}\num{5235}   & \small A small simple town with a mixture of residential and commercial buildings.               \\
                                                                   & Town04  & \num{5 } & \phantom{0}\num{3591}   & \small A small town embedded in the mountains with a special ``figure of 8''' infinite highway.  \\
                                                                   & Town06  & \num{11} & \phantom{0}\num{3980}   & \small Long many lane highways with many highway entrances and exits (with Michigan left).       \\
                                                                   & Town12  & \num{16} & \phantom{0}\num{9361}   & \small A large map, including high-rise, residential and rural environments.                     \\
      \midrule
                                                             total & 4       & \num{43} & \num{22167}  &                                                                                                  \\
      \midrule
     \multirow{1}{*}{\rotatebox[origin=c]{90}{\textit{val}}}       & Town10  & \num{11} & \phantom{0}\num{2825}   & \small A downtown area with skyscrapers, residential buildings and an ocean promenade.           \\
\midrule
           total                                                   & 1       & \num{11} & \phantom{0}\num{2825}   &                                                                                                  \\
\midrule
\midrule
           overall                                                 & \num{9} & \num{98} & \num{50045} \\
\bottomrule
\end{tabular}

}
\caption{
  Summary of our data splits, including CARLA~\cite{Dosovitskiy17} maps, number of randomly started simulation runs, and total number of frames for the given map.
  We include the short summary provided by CARLA for each map.
}
\vspace{-0.05in}
\label{tab:lisu_scenes}
\end{table*}

\begin{figure*}
    \centering
    \subfloat[Town02]
    {
        \includegraphics[width=0.48\linewidth]{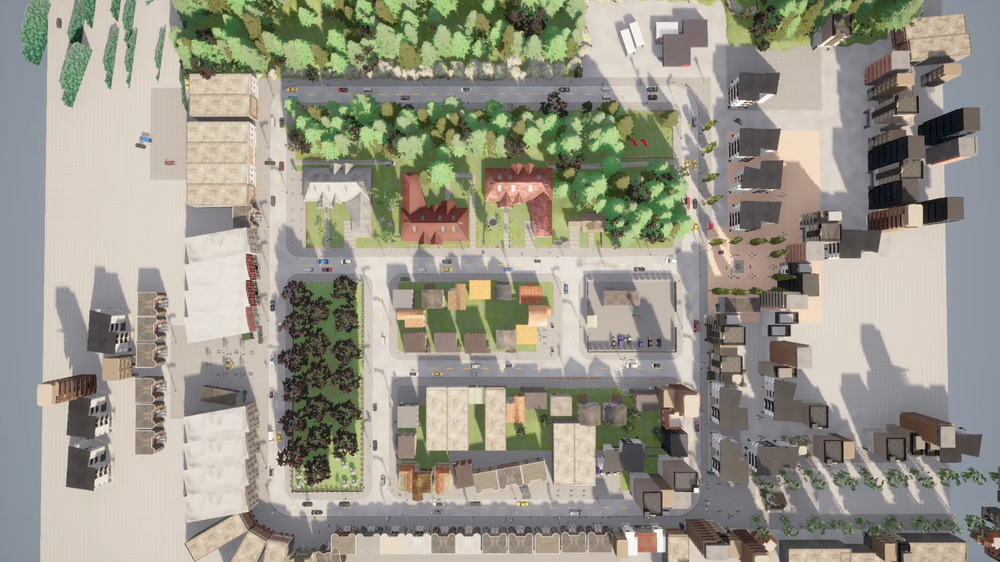}
        \label{fig:carla_maps_bev_town02}
    }
    \subfloat[Town03]
    {
        \includegraphics[width=0.48\linewidth]{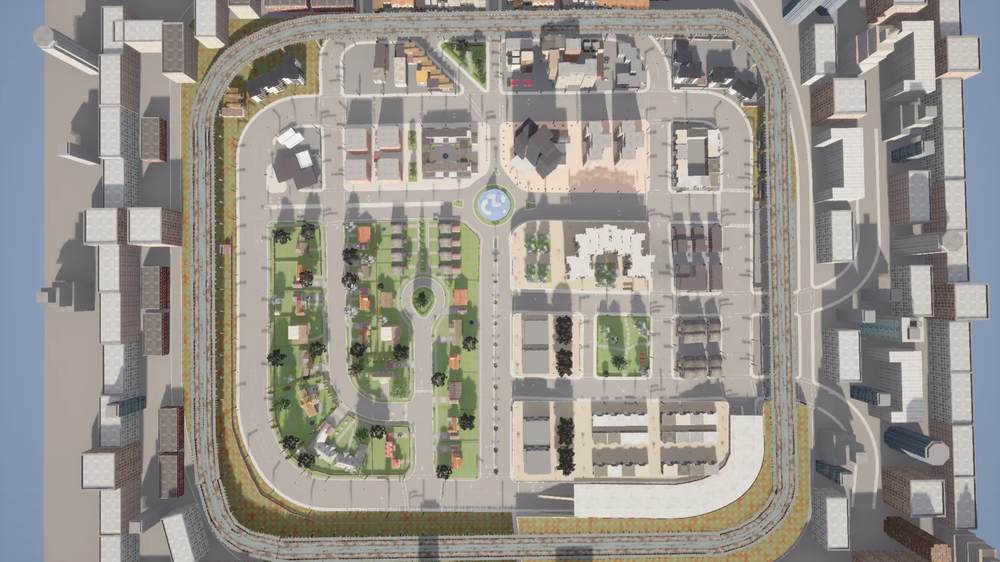}
        \label{fig:carla_maps_bev_town03}
    }
    \caption{
      Bird's eye view images of CARLA towns. Images taken from \url{https://carla.readthedocs.io/en/latest/core\_map/\#carla-maps}
    }
    \label{fig:carla_maps_bev}
\end{figure*}

\section{Implementation Details of Other Methods}
\label{sec:implementation_details_others}

We benchmarked our method against several state-of-the-art point cloud surface normal estimation methods: PCPNet~\cite{guerreroPCPNetLearningLocal2018}, GraphFit~\cite{liGraphFitLearningMultiscale2022}, Du \etal~\cite{duRethinkingApproximationError2023}, and SHS-Net~\cite{liSHSNetLearningSigned2023}.
These methods share a common training strategy, originally proposed in PCPNet, which involves randomly sampling query points and their $k$-nearest neighbors from each training sample.
While PCPNet's dataset consists of dense point clouds, our LiDAR data is significantly sparser. Consequently, we reduced the neighborhood size $k$ to $32$ to better adapt to the sparse nature of our data.
Furthermore, given our larger dataset, we decreased the number of training epochs to 10 for all methods.
All other hyperparameters were kept consistent with their original settings.

\section{Neural Surface Reconstruction}
\label{sec:abl_neural_surface_reconstruction}

Our experiments in Sec.~4.3 closely replicate existing benchmarks in neural surface reconstruction from LiDAR data~\cite{zhangReSimADZeroShot3D2024, guoStreetSurfExtendingMultiview2023}.
In these benchmarks, a mesh reconstructed from all available LiDAR data is queried with rays generated from the same data, and the resulting distances are compared to actual LiDAR measurements.
While this approach offers a convenient evaluation framework, it may not accurately reflect the method's true performance, as the same data is used both in training and evaluation.

Therefore, we propose a more rigorous evaluation protocol.
We randomly split LiDAR points from an entire sequence into two disjoint sets: a training and a testing set.
The training split is used exclusively in the training phase.
Subsequently, the reconstructed mesh is evaluated using the unseen testing set.
By ensuring that the training and testing sets are mutually exclusive, we can better identify model's potential limitations.

The proposed evaluation protocol proves particularly challenging for plain ReSimAD~\cite{zhangReSimADZeroShot3D2024}, as evident from the mesh reconstruction in \cref{fig:abl_nur_surf_rec} and LiDAR simulation in \cref{fig:nsr_lidar_sim}.
Limited training data hinders SDF generalization, leading to poor extrapolation in areas lacking ground truth signals, such as ridges in the mesh (\cref{fig:abl_nur_surf_rec}).
This noise propagates to the LiDAR simulation (\cref{fig:nsr_lidar_sim}), resulting in highly noisy point clouds.
Incorporating surface normals as an additional training signal mitigates these issues, leading to smoother meshes and consequently cleaner point clouds.
This improvement is also quantified in \cref{tab:sup_implicit_surface_reconstruction}

\begin{figure*}
    \centering
    \subfloat[1172406]
    {

      \resizebox{\textwidth}{!}{%
      \includegraphics{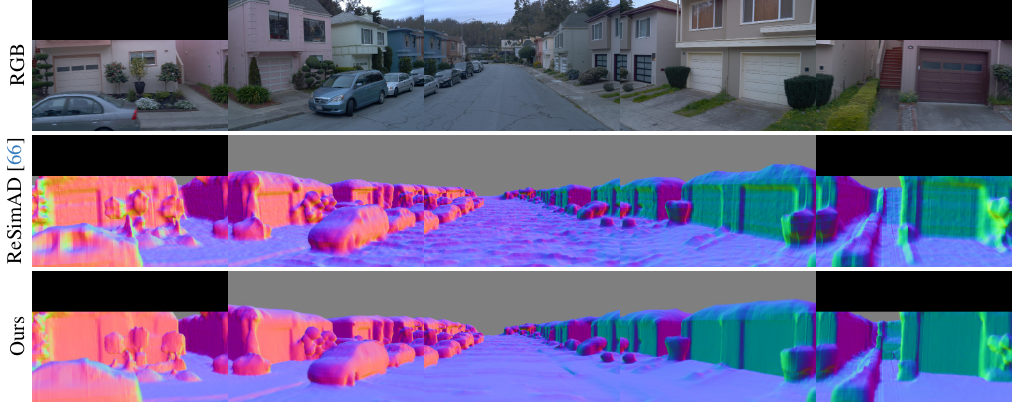}
      }
    }

    \subfloat[1522170]
    {
      \resizebox{\textwidth}{!}{%
      \includegraphics{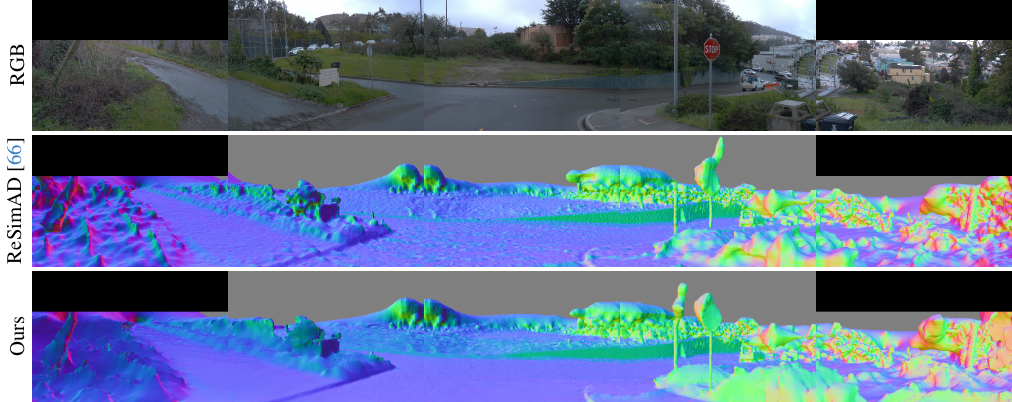}
      }
    }

    \subfloat[1006130]
    {
      \resizebox{\textwidth}{!}{%
      \includegraphics{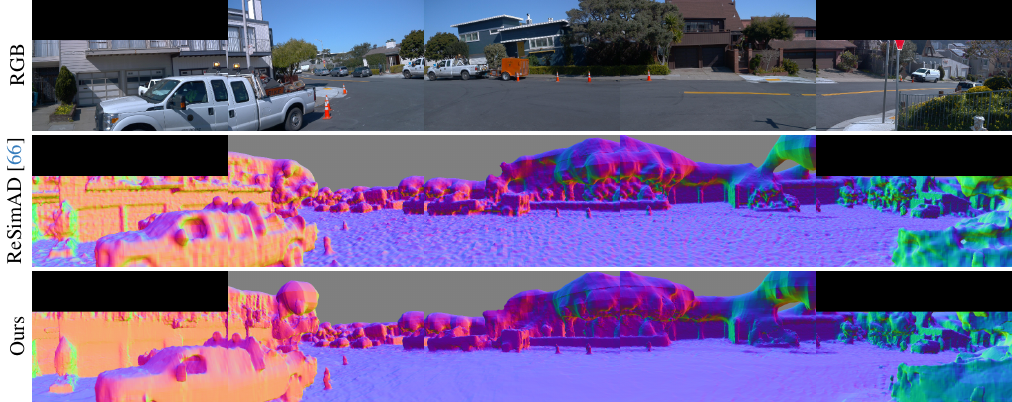}
      }
    }
    \caption{
      Mesh reconstruction results comparing plain ReSimAD~\cite{zhangReSimADZeroShot3D2024} and our method, which incorporates surface normals estimated from a model trained on our \dsname dataset and fine-tuned on Waymo Open Dataset~\cite{sunScalabilityPerceptionAutonomous2020}.
      Results are shown for three different Waymo sequences.
    }
    \label{fig:abl_nur_surf_rec}
\end{figure*}

\begin{figure*}
    \centering
    \includegraphics{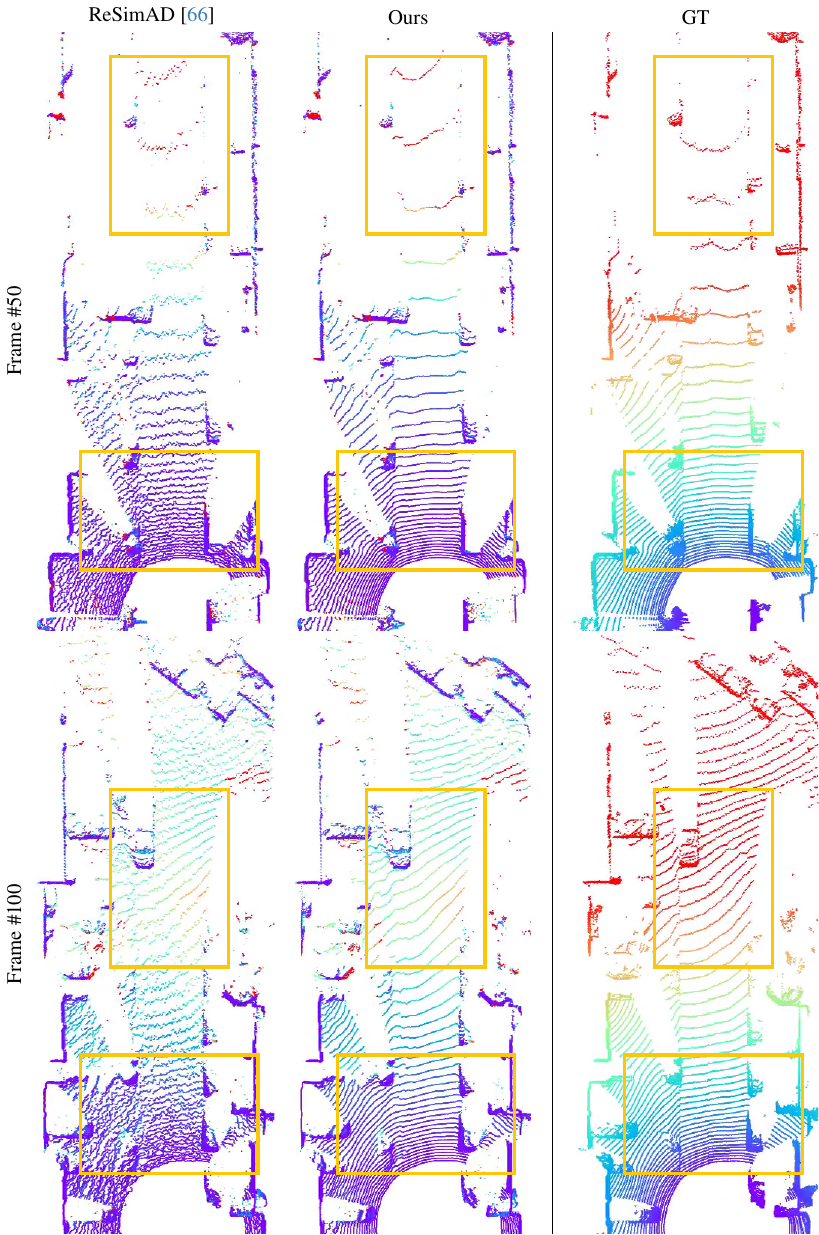}
    \caption{
      LiDAR simulation results comparing mesh reconstructed with plain ReSimAD~\cite{zhangReSimADZeroShot3D2024} (left) and our method, which leverages estimated surface normals from a model trained on our \dsname dataset and fine-tuned on Waymo Open Dataset~\cite{sunScalabilityPerceptionAutonomous2020} (middle).
      The color scale represents the deviation from ground truth distance, with blue indicating low error and red indicating high error.
      The rightmost image shows the ground truth LiDAR point cloud (used for mesh reconstruction), colored by distance from the sensor (blue: near, red: far).
    }
    \label{fig:nsr_lidar_sim}
\end{figure*}

\begin{table}
\centering
\begin{tabular}{ccc}
\toprule
\textbf{Seq.} & \textbf{ReSimAD}~\cite{zhangReSimADZeroShot3D2024} & \textbf{Ours}                         \\
\midrule
1027514       & 0.68~/~0.15                                        & \bestresult{0.65}~/~\bestresult{0.14} \\
1006130       & \bestresult{2.55}~/~\bestresult{0.23}              & 2.56~/~0.25                           \\
1137922       & 0.95~/~0.77                                        & \bestresult{0.92}~/~\bestresult{0.73} \\
1323841       & \bestresult{0.51}~/~\bestresult{0.11}              & \bestresult{0.51}~/~0.12              \\
1486973       & 0.83~/~\bestresult{0.16}                           & \bestresult{0.82}~/~0.17              \\
1522170       & 2.05~/~2.70                                        & \bestresult{1.78}~/~\bestresult{2.06} \\
1647019       & 1.01~/~0.70                                        & \bestresult{0.98}~/~\bestresult{0.63} \\
3425716       & 0.70~/~0.22                                        & \bestresult{0.68}~/~\bestresult{0.20} \\
9385013       & 0.52~/~0.20                                        & \bestresult{0.51}~/~\bestresult{0.18} \\
\midrule
average       & 1.09~/~0.58                                        & \bestresult{1.04}~/~\bestresult{0.50} \\
\bottomrule
\end{tabular}
\caption{
    We evaluate neural surface reconstruction on diverse Waymo sequences using Root Mean Square Error (RMSE)~/~Chamfer Distance (CD).
    Lower values indicate better performance.
    ReSimAD~\cite{zhangReSimADZeroShot3D2024} omits surface normal loss during reconstruction.
    Ours is a Waymo model, trained with our proposed self-training framework.
}
\vspace{-0.05in}
\label{tab:sup_implicit_surface_reconstruction}
\end{table}

\section{Qualitative Evaluation}
\label{sec:supp_qualitative}

We conduct a qualitative evaluation on the Waymo Open Dataset~\cite{sunScalabilityPerceptionAutonomous2020} to highlight the benefits of our \dsname.
By comparing SHS-Net~\cite{liSHSNetLearningSigned2023} trained on PCPNet~\cite{guerreroPCPNetLearningLocal2018} to our \dsname, we demonstrate the significant advantage of leveraging a dataset tailored to real-world LiDAR point clouds.
The lack of publicly available LiDAR datasets underscores the potential of our \dsname to advance the field, regardless of the specific method employed.
Notably, our self-supervised approach achieves impressive results when applied to a real-world dataset like Waymo, as visualized in \cref{fig:sup_qualitative}.

\begin{figure*}
    \centering
    \subfloat[SHS-Net~\cite{liSHSNetLearningSigned2023} trained on PCPNet~\cite{guerreroPCPNetLearningLocal2018}]
    {
      \includegraphics[width=0.8\linewidth,trim={0px 0px 0 170px}, clip]{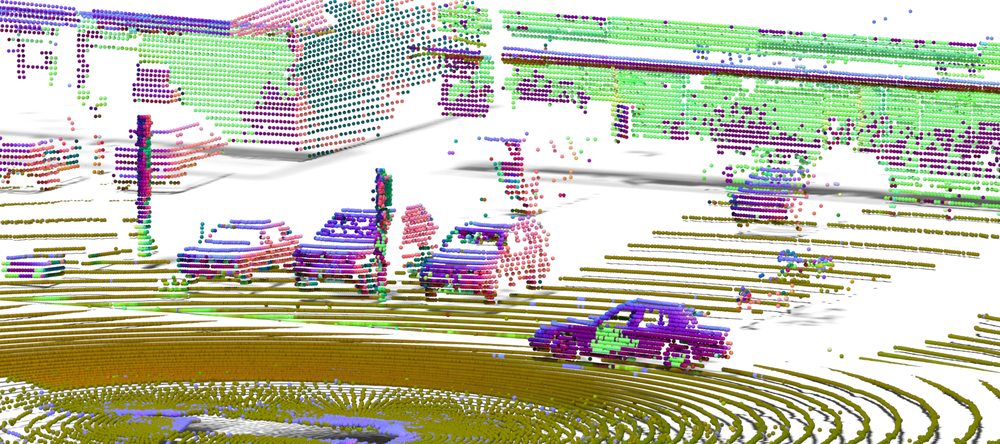}
      \label{fig:sup_qualitative:shs_pcp}
    } \hfill
    \subfloat[SHS-Net~\cite{liSHSNetLearningSigned2023} trained on our \dsname]
    {
      \includegraphics[width=0.8\linewidth,trim={0px 0px 0 170px}, clip]{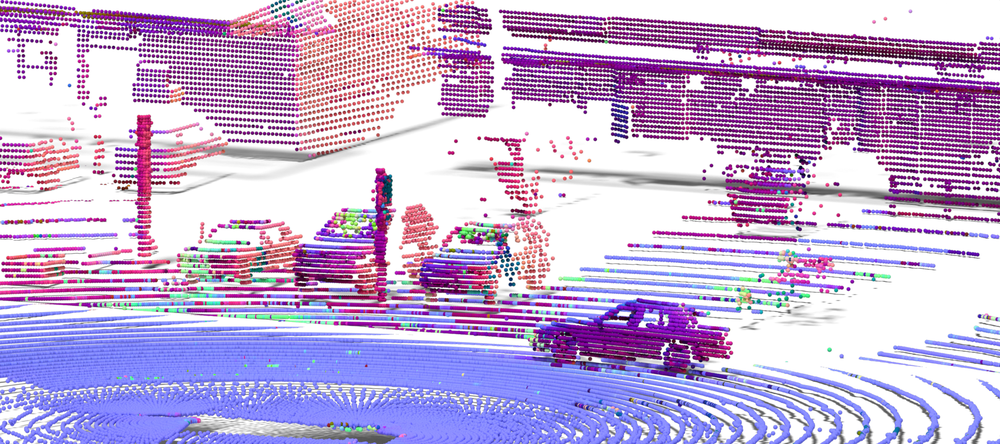}
      \label{fig:sup_qualitative:shs_lisu}
    } \hfill
    \subfloat[Direct transfer with our method]
    {
      \includegraphics[width=0.8\linewidth,trim={0px 0px 0 170px}, clip]{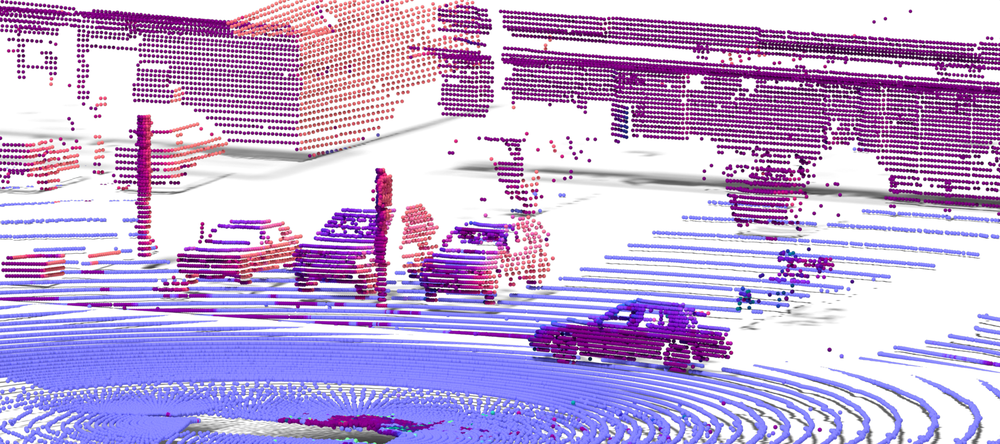}
      \label{fig:sup_qualitative:ours}
    } \hfill
    \subfloat[Unsupervised domain adaptation from \dsname to Waymo with our method]
    {
      \includegraphics[width=0.8\linewidth,trim={0px 0px 0 170px}, clip]{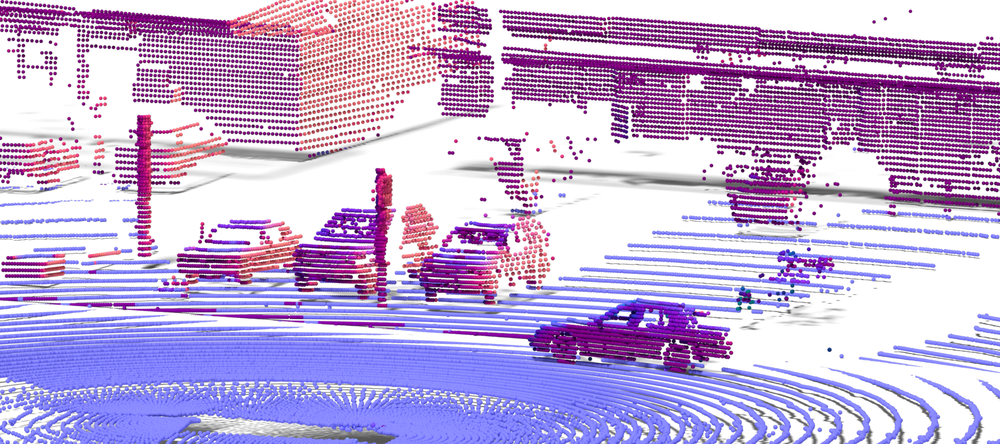}
      \label{fig:sup_qualitative:ours_da}
    }
    \caption{
      Qualitative comparison of SHS-Net~\cite{liSHSNetLearningSigned2023} directly transferred from PCPNet~\cite{guerreroPCPNetLearningLocal2018} \hyperref[fig:sup_qualitative:shs_pcp]{(a)} and our \dsname dataset \hyperref[fig:sup_qualitative:shs_lisu]{(b)}.
      Our dataset, tailored for LiDAR surface normal estimation, yields superior results.
      Additionally, we demonstrate the effectiveness of our method for both direct transfer \hyperref[fig:sup_qualitative:ours]{(c)} and self-supervised domain adaptation \hyperref[fig:sup_qualitative:ours_da]{(d)} on a challenging Waymo Open Dataset frame.
    }
    \label{fig:sup_qualitative}
\end{figure*}

\section{Loss-Regularization Trade-off}
\label{sec:loss_reg_tradeoff}

In the following section, we present an ablation study to substantiate our selection of the $\gamma$ parameter (Eq.~(7) of the main manuscript), which balances loss minimization and regularization.
To expedite the training and evaluation phases, we utilized $50\%$ and $20\%$ subsets of the original training and evaluation data, respectively.

The hyperparameter $\gamma$ balances the trade-off between noise and smoothness in the final predictions.
Lower values of $\gamma$ encourage the model to prioritize edge preservation, potentially leading to noisier predictions.
Conversely, higher values of $\gamma$ promote smoother predictions, which may result in the loss of fine-grained details and edge information.
In the extreme case, when $\gamma=1$, the model's predictions become almost entirely smooth, with minimal edge detection.
In our experiments, we found that $\gamma=0.1$ provided an optimal balance between noise and detail, as illustrated in \cref{fig:gamma_abl}.

\begin{figure}
\centering
\resizebox{\columnwidth}{!}{%
\includegraphics{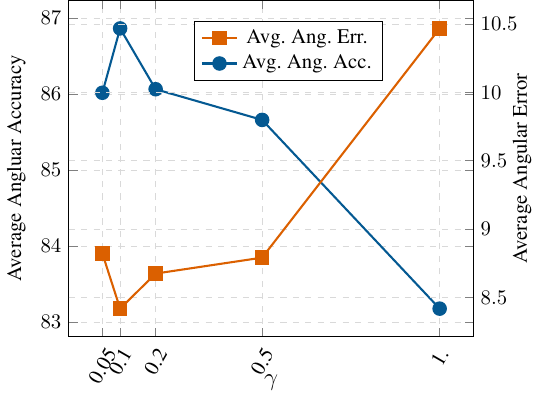}
}%
\caption{
  Average angular accuracy ($\uparrow$) computed across thresholds $\{5.0^\circ, 7.5^\circ, 11.25^\circ, 22.5^\circ, 30.0^\circ\}$ and average angular error ($\downarrow$) calculated across mean, median and RMSE, for varying $\gamma$.
}
\label{fig:gamma_abl}
\end{figure}

\section{\texorpdfstring{$k$ Neighborhood Graph Ablation}{k Neighborhood Graph Ablation}}
\label{sec:k_nbhood_graph_abl}

To construct the $k$-neighborhood graph $\mathcal{G}$ for both Spatial Graph Total Variation (SGTV) and Temporal Graph Total Variation (TGTV) (Sec.~3.2), we require a hyperparameter $k$ to specify the size of the local neighborhood.
In the following experiments, we fix $\gamma=0.1$ and systematically vary $k$ to assess its influence on the model's overall performance.
To expedite the training and evaluation phases, we utilized $50\%$ and $20\%$ subsets of the original training and evaluation data, respectively.

The parameter $k$ controls the model's output smoothness, with higher values leading to increased smoothing and potential loss of detail (\eg $k=32$ in \cref{fig:abl_k}).
Conversely, smaller values of $k$ (\eg $4$) may not provide sufficient smoothing, resulting in noisy outputs.
As \cref{fig:abl_k} illustrates, $k=8$ offers a favorable balance between smoothness and detail preservation.
As a general guideline, we recommend setting $k$ to the median number of points within a $0.1$-meter radius around each point in the input data.

\begin{figure}
\centering
\resizebox{\columnwidth}{!}{%
\includegraphics{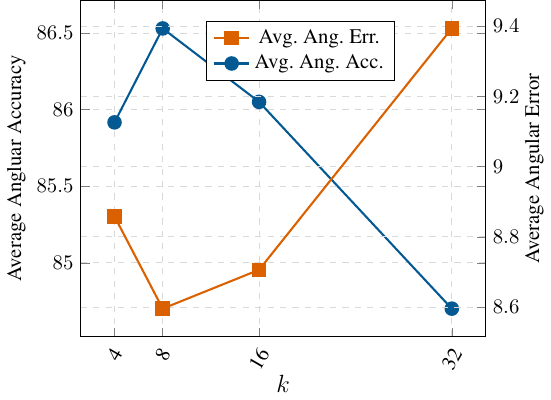}
}%
\caption{
  Average angular accuracy ($\uparrow$) computed across thresholds $\{5.0^\circ, 7.5^\circ, 11.25^\circ, 22.5^\circ, 30.0^\circ\}$ and average angular error ($\downarrow$) calculated across mean, median and RMSE, for varying $k$.
}
\label{fig:abl_k}
\end{figure}

\section{Weighted Adjacency Matrix Ablation}
\label{sec:gamma_ablation}

Edge weights in our graph $\mathcal{G}$ for both SGTV and TGTV (Sec.~3.2 of the main manuscript) are computed using an exponential decay function,
\begin{align}
  w(x) = \exp \left( -\frac{x^2}{\sigma^2} \right) \text{,}
\end{align}
where $x$ represents the Euclidean distance between two graph nodes (\ie points).
The decay constant $\sigma$ determines the rate at which the edge weight decreases with increasing distance.
We depict the influence of different $\sigma$ in \cref{fig:sigma_different}.

\begin{figure}
\centering
\resizebox{\columnwidth}{!}{%
\includegraphics{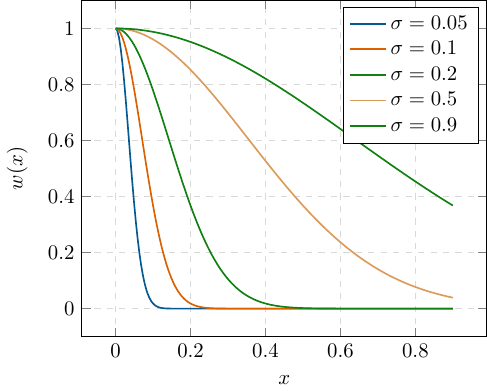}
}%
\caption{
  Effect of the decay constant $\sigma$ on edge weights for different distances $x$ in meters.
}
\label{fig:sigma_different}
\end{figure}

To study the impact of the hyperparameter $\sigma$, we conducted an ablation study with $\gamma=0.1$ and $k=8$, varying $\sigma$ across multiple runs.
For efficiency, we used 50\% of the training data and 20\% of the evaluation data.
Results showed that smaller $\sigma$ values produce sparse graphs, reducing regularization, while larger values introduce noise by connecting distant points, potentially belonging to different surfaces (\eg $0.2$ meters apart).
Empirically, $\sigma=0.1$ was found optimal, as shown in Figure~\ref{fig:abl_sigma}.

\begin{figure}
\centering
\resizebox{\columnwidth}{!}{%
\includegraphics{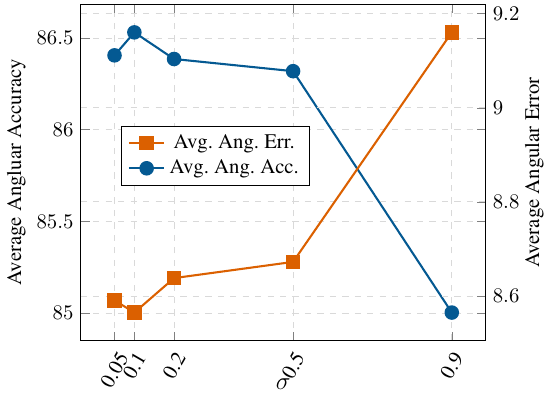}
}%
\caption{
  Average angular accuracy ($\uparrow$) computed across thresholds $\{5.0^\circ, 7.5^\circ, 11.25^\circ, 22.5^\circ, 30.0^\circ\}$ and average angular error ($\downarrow$) calculated across mean, median and RMSE, for varying $\sigma$.
}
\label{fig:abl_sigma}
\end{figure}

 \fi

\end{document}